# Predicting gene essentiality and drug response from perturbation screens in preclinical cancer models with LEAP: Layered Ensemble of Autoencoders and Predictors

Barbara Bodinier*#, Gaetan Dissez*, Lucile Ter-Minassian, Linus Bleinstein, Roberta Codato, John Klein, Eric Durand, Antonin Dauvin*

* These authors contributed equally
# Corresponding author

## Abstract

High-throughput preclinical perturbation screens, where the effects of genetic, chemical, or environmental perturbations are systematically tested on disease models, hold significant promise for machine learning-enhanced drug discovery due to their scale and causal nature. Predictive models trained on such datasets can be used to (i) infer perturbation response for previously untested disease models, and (ii) characterise the biological context that affects perturbation response.

Existing predictive models suffer from limited reproducibility, generalisability and interpretability. To address these issues, we introduce a framework of Layered Ensemble of Autoencoders and Predictors (LEAP), a general and flexible ensemble strategy to aggregate predictions from multiple regressors trained using diverse gene expression representation models. LEAP consistently improves prediction performances in unscreened cell lines across modelling strategies.

In particular, LEAP applied to perturbation-specific LASSO regressors (PS-LASSO) provides a favorable balance between near state-of-the-art performance and low computation time. We also propose an interpretability approach combining model distillation and stability selection to identify important biological pathways for perturbation response prediction in LEAP. Our models have the potential to accelerate the drug discovery pipeline by guiding the prioritisation of preclinical experiments and providing insights into the biological mechanisms involved in perturbation response. The code and datasets used in this work are publicly available.

# Introduction

Gene and drug perturbation screens have transformed preclinical studies by enabling the systematic evaluation of a vast number of perturbations across diverse biological contexts (e.g. different cancer cell lines), helping the identification and positioning of novel therapeutic targets and compounds [1–4]. These datasets are particularly valuable as the perturbations—whether genetic or chemical—are applied in a controlled experimental setting, allowing for direct observation of cause-and-effect relationships.

In oncology, significant initiatives have been undertaken to create and publicly share datasets of unprecedented scale. The Cancer Dependency Map (DepMap) project [5], for instance, contains data on the essentiality of over 17,000 genes in more than 1,000 cancer cell lines obtained from high-throughput CRISPR-Cas9 viability screens. Gene essentiality captures the change in cell viability following targeted CRISPR-Cas9 knockout. Essential genes are critical for cell survival and proliferation in specific molecular contexts. Similarly, studies like GDSC[6], CCLE[7], CTRP[8], and PRISM[9] have measured the impact of various compounds on cancer cell line viability. Drug response experiments involve exposing cancer cell lines to varying concentrations of compounds and quantifying cell viability. These studies also gather deep molecular characterization of the cell lines, including RNA sequencing to define the biological context. This enables the investigation of how molecular context influences cancer vulnerability and paves the way for more personalized medicine.

In this paper, we train machine learning models to predict viability responses to perturbations from pre-perturbation molecular profiles of the disease models in which they were tested. These approaches can be used to predict the impact of a perturbation in untested disease models or patients[10,11], thus providing insights into the relevance of targets or compounds in different indications, disease models, or even patients. This capability can aid in prioritizing targets, compounds, or preclinical experiments in the early stages of the drug discovery pipeline.

Previous studies have proposed machine learning approaches for the prediction of gene essentiality [10–12] or drug response with promising performances [13–18]. Existing approaches can be classified into two categories: perturbation-specific (PS) and pan-perturbation (PP) models. Perturbation-specific models [11,13–15] aim to predict the response of different biological systems to a single perturbation, using molecular data as input. When data includes multiple perturbations, separate models are trained for each one. In contrast, pan-perturbation models [10,12,16–18] are trained on data from multiple perturbations simultaneously. In this paper, a pan-perturbation model is a single-label regressor [10,12,16–18] trained on all available pairs of disease models and perturbations. A PP model leverages features from both the disease model (molecular data) and the perturbations, which may be represented by pathway-based fingerprints for a genetic perturbation or by the molecular structure for a drug. Note that a multi-label regressor predicting the response to multiple perturbations from the molecular profiles of the disease models may also be considered as a pan-perturbation model, but such models are omitted from the present paper as they have not been studied as extensively. Pan-perturbation models have the potential to uncover both mechanisms that are specific to a perturbation and those that are shared across perturbations. Unlike perturbation-specific models, pan-perturbation models leveraging perturbation descriptors can intrinsically predict responses to untested perturbations, provided these can be appropriately captured by their embeddings. However, in this paper, we focus on predictions of responses to seen perturbations in unscreened disease models, such as new cell lines, tissues, or more complex disease models like patient-derived xenografts (PDX). This ability to infer perturbation responses in diverse disease models remains a key need in drug discovery.

Existing perturbation-specific models have been trained using a variety of algorithms including both linear (e.g. Ridge, LASSO, Elastic Net) and non-linear (e.g. neural networks, XGBoost, Random Forests) regressors[19]. Pan-perturbation methods such as DeepDep[10] for gene essentiality prediction, or tDNN[18] and DrugCell[17] for drug response prediction combine representations of disease models and perturbations. Previous work from our team, OmicsRPZ[12], improved gene essentiality predictions by first generating these two separate representations and then training a Light Gradient Boosting Model (LGBM) on these combined features. We leverage our previous work on representation learning, but focus here on the choice of a prediction model and the development of a novel ensemble framework.

We propose Layered Ensemble of Autoencoders and Predictors (LEAP), a novel and general ensemble approach that can be applied to both perturbation-specific or pan-perturbation models to predict perturbation response using gene expression data. LEAP uses an ensemble of regressors trained in randomized subsets of cell lines embedded with different representation models. We apply the LEAP framework using representations obtained from Masked Autoencoders (MAE) that are randomly initialized with different seeds, as they were previously identified as the best gene expression representation models for such tasks in previous work[12]. We pre-train the autoencoders using unlabelled data on various disease models (cell lines and PDX) to capture common patterns across biological contexts and reduce computational time. LEAP leverages (i) recent advances in deep representation learning, which have achieved state-of-the-art results in various domains, and (ii) model ensembling, which enhances prediction precision and robustness by utilizing model randomized diversity [20–22].

To facilitate the biological interpretation of LEAP, we also introduce a pathway-based explanation framework that overcomes the challenges posed by the high-dimensionality and strong correlations patterns that are inherent to gene expression data. For this, we create Reactome[23] pathway-level features constructed from linear combinations of genes that belong to the same pathway[24]. We then distill LEAP by supervising the training of LASSO regressors using these pathway-level features as input[11]. Stability selection[25] then enables the identification of stable sets of biological pathways that drive LEAP predictions.

We evaluate the performance of LEAP for gene essentiality and drug response prediction in unscreened cell lines, tissues or disease models by adapting the setup previously used in the DeepDep[10] and OmicsRPZ[12] studies. We use more recent releases of the datasets, and we standardize the training and evaluation framework between gene essentiality and drug response tasks. We quantify the increase in prediction performance by using LEAP with multiple modeling strategies, including perturbation-specific and pan-perturbation approaches with both linear and non-linear regressors. In line with previous publications[15,26–29], we consider that drug discovery operates within a transductive framework, which implies that unlabelled test data (e.g. RNAseq profiles) are available during training. This can be leveraged by pretraining the representation model on both labelled and unlabelled data with limited impact on the prediction performances, as demonstrated in our ablation studies. In this paper, we pre-train MAE representation models on all available RNAseq data and compare prediction performances obtained using different regression approaches on the same data representation. To evaluate the ability of our interpretability strategy to recover important biological pathways for prediction, we compare prediction performances obtained with LEAP and with a LASSO regressor using selected pathway-level features only.

We show that LEAP improves performances in predicting perturbation response in unlabeled biological contexts regardless of the modeling strategy. Notably, we show in an ablation study that the ensemble of different data representations in LEAP outperforms methods that ensemble prediction models alone. We identify that LEAP applied to perturbation-specific LASSO regressors offers a good balance between near state-of-the-art performance and low computational burden. Human evaluation for the top predicted compounds confirms plausible links between some of the pathways driving predictions in LEAP and known mechanisms of actions for these drugs.

# Results

## Outline

We propose LEAP, a novel approach to predict perturbation response (gene essentiality or drug response) in a disease model based on gene expression before perturbation (Figure 1). We use publicly available data from perturbation experiments conducted in cell lines in the DepMap[5] and PharmacoDB[30] studies, and in Patient-Derived Xenografts (PDX) in the PDX Encyclopedia[31] (Supplementary Table 1, Supplementary Figure 1). We consider three challenges, where we aim to predict perturbation response in (i) unscreened cell lines, (ii) cell lines from an unscreened tissue of origin, or (iii) unscreened PDX (Table 1).

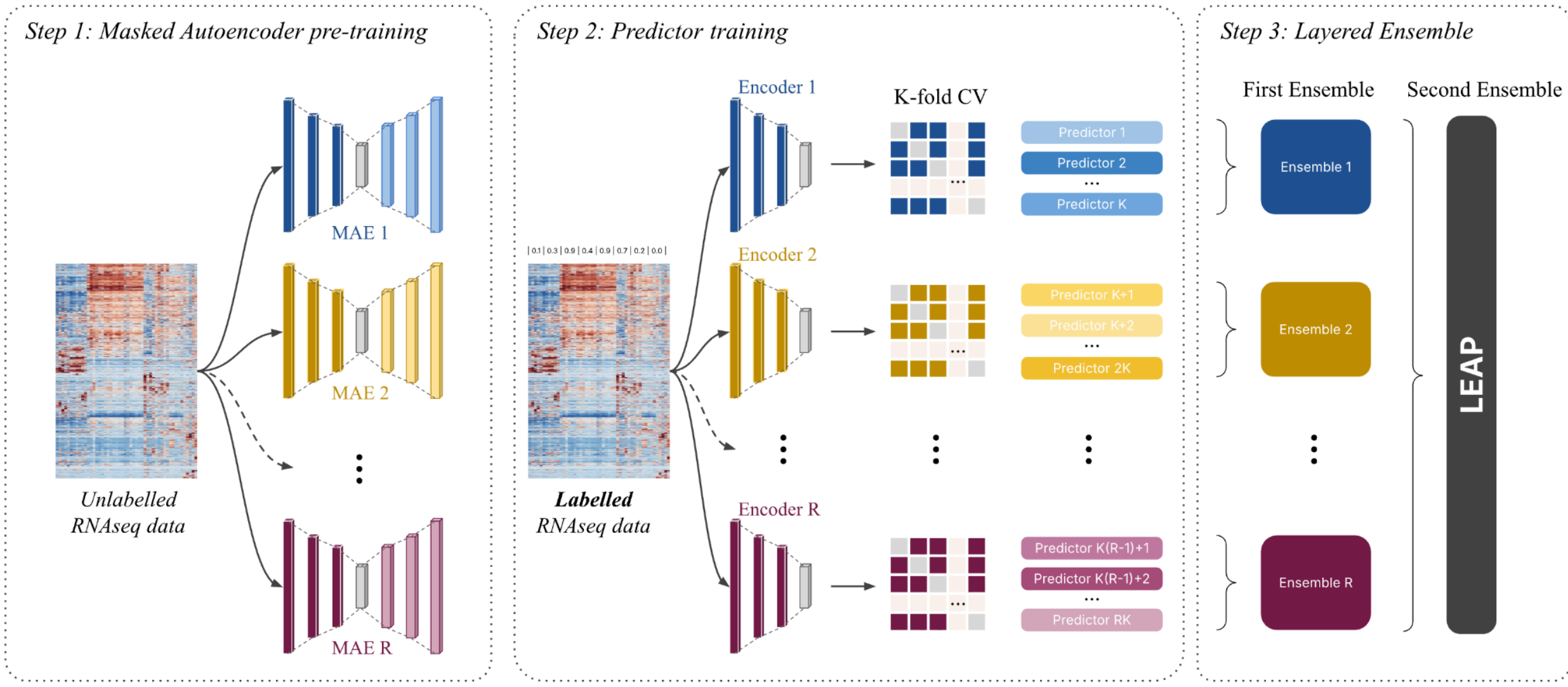

***Figure 1. Illustration of LEAP.*** *The training data is first transformed using five MAE models that have been trained with different random seeds. We use a 5-fold cross validation on each of the five representations of the data and ensemble the five regressors trained on the different cross validation folds with the best performing hyper-parameters. Thus, 25 regression models form the final LEAP model.*

| Challenge | Task id | Label | Data | Number of cell lines | Number of perturbations | Number of (cell line, perturbation ) pairs |
|---|---|---|---|---|---|---|
| i - Prediction in unscreened cell lines | 1 | Gene dependency | DepMap (version 23Q4) | 1,019 | 1,539 | 1,568,222 |
| | 1’ | Gene dependency | DepMap (OmicsRPZ[12]) | 893 | 1,223 | 1,092,139 |
| | 2 | Drug response (AAC) | PharmacoDB | 1,399 | 649 | 551,024 |
| ii - Prediction in unscreened tissue of origin | 3 | Gene dependency | DepMap (version 23Q4) | 1,019 | 1,539 | 1,568,222 |
| | 4 | Drug response (AAC) | PharmacoDB | 1,399 | 580 | 528,463 |
| iii - Prediction in unscreened disease model | 5 | Drug response (AAC) | PharmacoDB, PDX Encyclopedia | 1,390 (140 PDX) | 5 | 6,129 (227 PDX) |

***Table 1. Task description.*** *For each of our six tasks, we report the type of label to predict, dataset version, number of cell lines, number of perturbations and number of pairs of cell lines and perturbations. Task 1’ is a similar task to task 1, using an older version of the data and a different set of perturbations in order to compare our approach with previous work.*

We propose a modelling strategy that builds upon our previous pan-perturbation approach OmicsRPZ[12], which proposed SOTA models to predict gene essentiality alongside a benchmark strategy. We use the Masked Auto-Encoder (MAE) identified in this previous work as one of the best approaches across tasks. We apply the LEAP framework to different prediction algorithms: three pan-perturbation (PP) models, including PP-LGBM, which is an update of the best model from OmicsRPZ using our MAE model, PP-tDNN[18] and PP-MLP, and four perturbation-specific (PS) models, including the baseline PS-KNN, PS-MLP, PS-LGBM and PS-LASSO.

In LEAP, we design a two-layer ensembling strategy aimed at enhancing the robustness of both the data representation and the prediction model (Figure 1). We train five MAE models with different random initializations on the same large dataset combining RNAseq data from all available cell lines and PDX samples. We then train regression models on the five different data representations generated by the MAEs. We tune the hyper-parameters of each regressor by maximizing Spearman's correlation using a 5-fold cross-validation within the training set. For the first layer of ensembling, we average the predictions from the five regression models that are trained for hyper-parameter tuning with the best performing hyperparameters. For the second layer of ensembling, we average the predictions of the five sets of regressors obtained with different data representations. In total, LEAP ensembles 25 regression models. In our approach, the randomization of the initial seeds of the autoencoder generates embeddings that capture complementary signals from RNAseq data. The second layer of ensembling leverages these complementary signals and yields better prediction performance. We quantify the change in prediction performance when applying LEAP to each of these prediction algorithms compared to base algorithms with the same hyper-parameter tuning and only the first layer of ensembling. We further compare our approach with published models, including DeepDep[10] and all models trained in the OmicsRPZ[12] study on the gene dependency prediction in unlabeled cell lines (challenge i, task 1').

Finally, we extract the most important biological pathways to predict perturbation response in LEAP. For this, we create an expression dataset of lower dimensionality with 1,832 Reactome pathway features, each defined as the first principal component of a PCA applied on all genes in the pathway. We then train LASSO regressors using pathway-level features to predict labels inferred by LEAP and use stability selection to identify important pathways.

## LEAP consistently improves perturbation response predictions compared to the base model

Applying LEAP consistently yields an increase in prediction performance compared to the base algorithm for all modeling approaches in tasks 1 and 2 (increase ranging from 1.4% for PS-KNN in task 2 to 4.4% for PP-tDNN in task 1, Figure 2A-B, Supplementary Table 2). In particular, LEAP applied to PS-LASSO (Spearman's correlations of 0.321 in task 1 and 0.350 in task 2), PS-MLP (0.307 in task 1 and 0.346 in task 2) and PP-LGBM (0.309 in task 1 and 0.367 in task 2) regressors stand out as the best performing approaches. LEAP (PS-LASSO) provides the best balance between prediction performance and computation time (approximately 30x and 150x faster than PS-MLP and PP-LGBM respectively on task 1, 45x and 340x faster on task 2, Supplementary Figure 2).

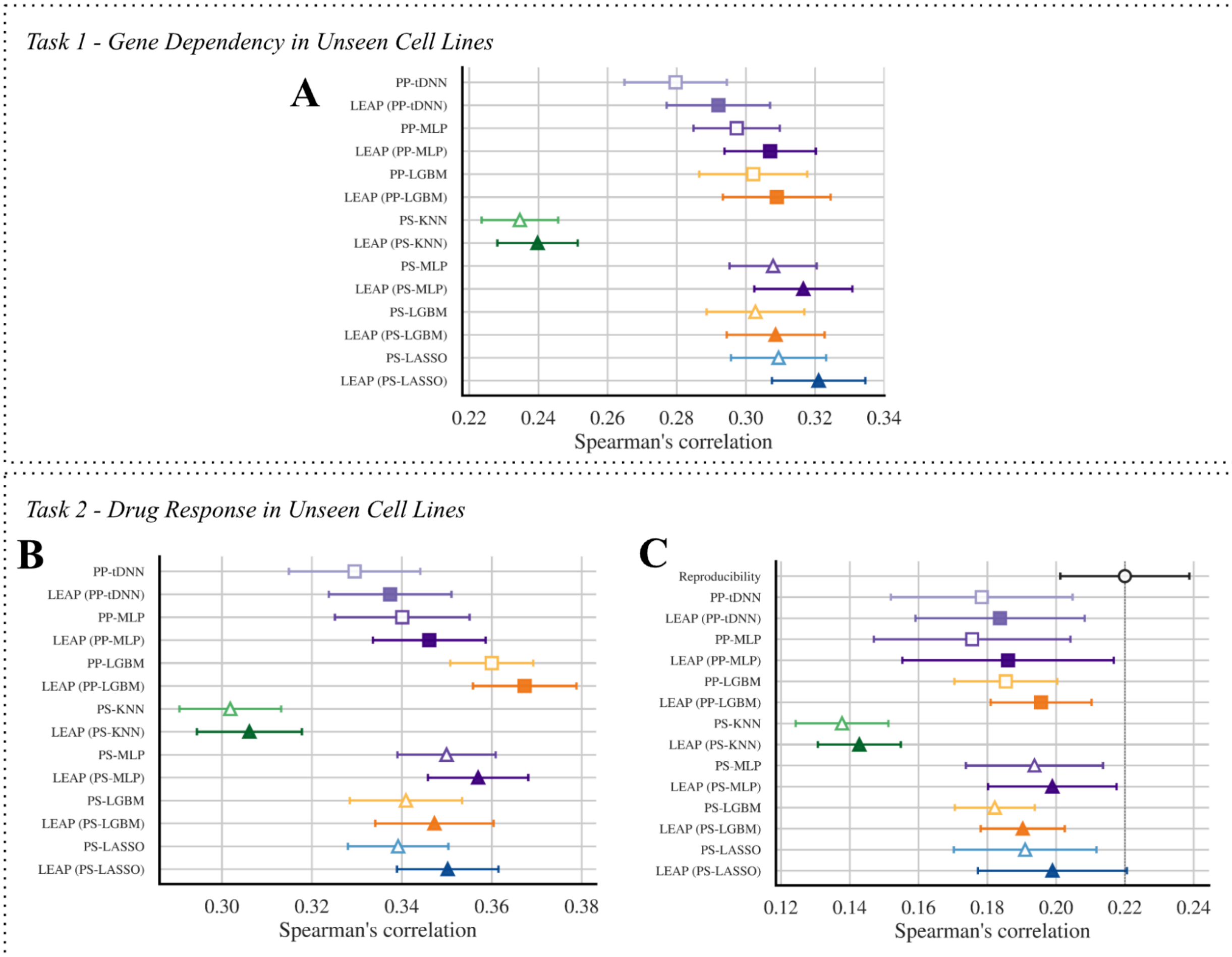

***Figure 2. Performances in predicting perturbation response in unscreened cell lines.*** *We compare the performances of our novel approach (LEAP, filled data points) with base models (empty data points). Base models only include the first layer of ensembling, where predictions obtained from regressors trained in each the 5 CV folds are averaged. We evaluate the performances in predicting* ***A*** *gene dependency in unscreened cell lines,* ***B*** *drug response in unscreened cell lines, and* ***C*** *drug response in unscreened cell lines from an external validation study (PRISM). Performances are evaluated over 10 repeated holdout splits (****A****,* ***B****) or over the entire external validation dataset for models trained on the 10 same training sets (****C****).*

LEAP (PS-LASSO) generates better performances than a baseline approach using target expression only in task 1, which confirms that this model can capture more complex predictive predictive signals from multiple genes (Supplementary Figure 3). LEAP (PS-LASSO) also outperforms all other models in the external validation study (Figure 2C). The prediction performances obtained with LEAP (PS-LASSO) are close to the drug response measurement reproducibility across studies (Supplementary Figure 4). The variability in prediction performance across perturbations is correlated with the standard deviation of the labels in both tasks 1 and 2 (Supplementary Figure 5).

LEAP (PS-LASSO) and LEAP (PS-MLP) also generalise better to unscreened tissues in tasks 3 and 4 (Supplementary Figure 6). We observe an increase in prediction performance when using LEAP on top of PS-LASSO approaches for all tissues (Supplementary Figure 7). LEAP also yields an increase in performance over PS-LASSO models for 3 of the 5 drugs that are available in our task 5 in PDX (Supplementary Figure 8, Supplementary Table 3). Taken together, these results demonstrate that our LEAP framework improves performances regardless of the choice of prediction heads.

## The ensembling framework improves performances in LEAP

We evaluate the impact of each of our methodological choices on the prediction performance in an ablation study (Supplementary Figure 9, Supplementary Table 4). We observe a small drop in prediction performance when pre-training the MAE representation model in all available DepMap cell lines instead of cell lines used in the training set only. Pre-training the MAE representation models on (i) DepMap only, (ii) DepMap and GDSC, or (iii) DepMap, GDSC and PDXE only yields limited changes in prediction performances.

We distinctly evaluate the gain in performance due to the first and second layers of our ensemble learning strategy in this ablation study (Supplementary Table 4, Supplementary Figure 9). The ensembling of the five regressors trained on the splits of the 5-fold cross validation (first layer ensembling) yields an increase in performance of 1% (0.306 to 0.309 and 0.335 to 0.339 respectively in tasks 1 and 2) compared to the use of a single regressor trained on the full training set. Further ensembling the 25 regressors trained on 5 different data representations (second layer ensembling) yields a larger 3 to 4% (0.309 to 0.321 and 0.339 to 0.350 respectively in tasks 1 and 2) increase in performance (Supplementary Figures 9 and 10).

To further assess the relevance of the second layer ensembling, we compare the performance of LEAP with an ensemble of 25 LASSO regressions trained on different subsets of the same deep representation (Supplementary Table 5). The ensembling of 25 regressors trained on the same data representation yields a limited increase (between 1 and 2% on tasks 1 and 2) in performance compared to the ensembling of LEAP. This suggests that the larger gain in performance in LEAP is due to the use of multiple data representations in the second layer ensembling and not only to the larger number of regressors that are ensembled.

## LEAP (PS-LASSO) outperforms state-of-the-art approaches in unscreened cell lines

We compare LEAP (PS-LASSO) to previous approaches for the prediction of gene essentiality in cell lines using the same data and evaluation framework as in OmicsRPZ[12]. LEAP (PS-LASSO) outperforms all approaches proposed in OmicsRPZ with different representation models and yields an increase in average per-perturbation Spearman's correlation of 16% (0.256 to 0.297) compared to the best OmicsRPZ model (Figure 3, Supplementary Table 6).

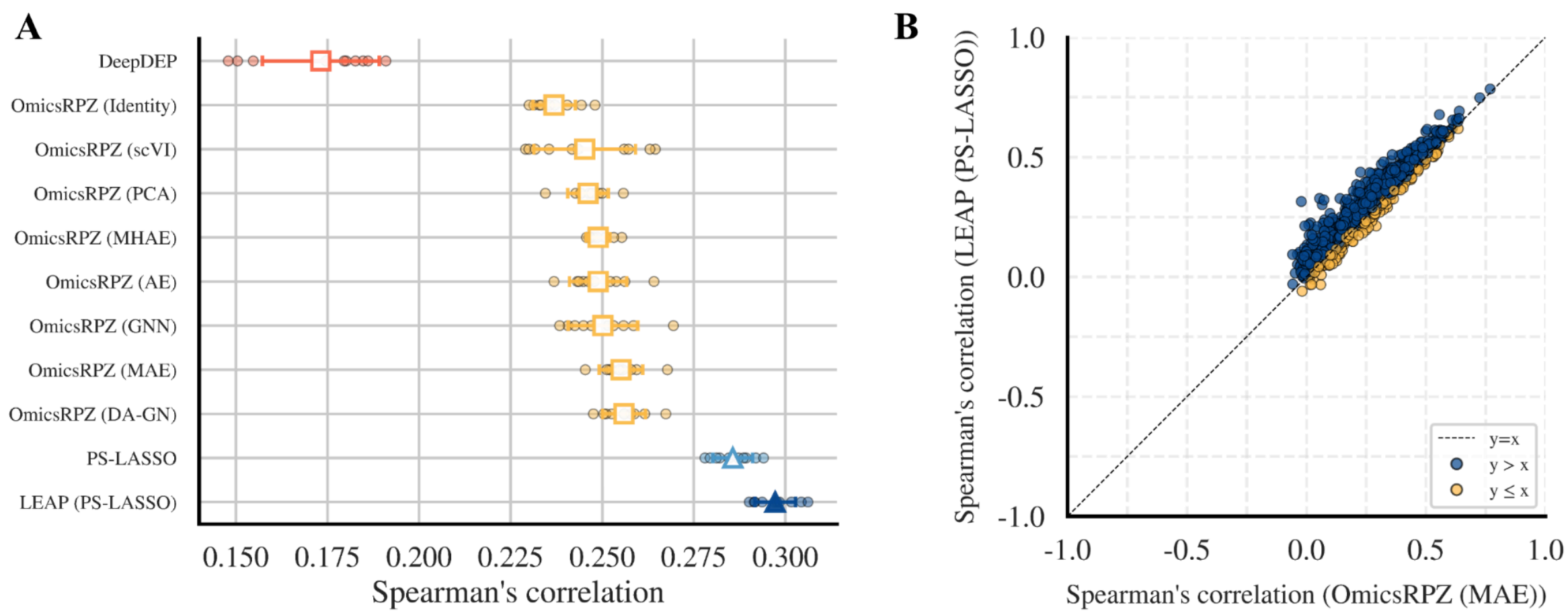

***Figure 3. Comparison of the performances in predicting gene dependency in unscreened cell lines using LEAP (PS-LASSO) or existing approaches (task 1').*** *We compare the performance of our approach LEAP (PS-LASSO) with previously reported performances obtained in DeepDEP[10] and using different representation models in OmicsRPZ[12]. Models are evaluated for the 1,223 gene perturbations included in the OmicsRPZ study over 10 repeated holdout test sets of unscreened cell lines. (A) Boxplots for different models where each point indicates the average Spearman's correlation coefficient over the 1,223 gene perturbations obtained in each of the 10 test sets. (B) Per-perturbation performances obtained using the best model from OmicsRPZ (x-axis) or using our model (y-axis). We report the average Spearman's correlation of each perturbation over the 10 test sets. Note that the performances of OmicsRPZ model are not recomputed and were obtained with different training-test splits of the same dataset.*

# LEAP's predictions are driven by stable, biologically relevant pathway features

To enhance the biological interpretability of LEAP, we develop a pathway-based explanation framework that distills LEAP into stable pathway signatures. For each dependency or drug, we identify a subset of Reactome pathways predictive of LEAP outcomes (PS-LASSO) using a stability selection[25] approach applied to pathway-level expression profiles. This results in perturbation-specific lists of biological pathways. Our distillation method generates stable results, with stability scores around 0.81 for task 1 and 0.82 for task 2 (Supplementary Figure 11, Supplementary Tables 7 and 8). We observe strong correlations between the prediction performances obtained using (i) selected pathways only as input in a LASSO regressor, or (ii) LEAP (PS-LASSO) for both tasks 1 and 2 (Figure 4). We report an average decrease in prediction performance of 0.039 in task 1 and 0.030 in task 2 when using selected pathways only compared to LEAP (PS-LASSO). These results suggest that our interpretability strategy successfully pinpoints the most important biological pathways to predict perturbation response.

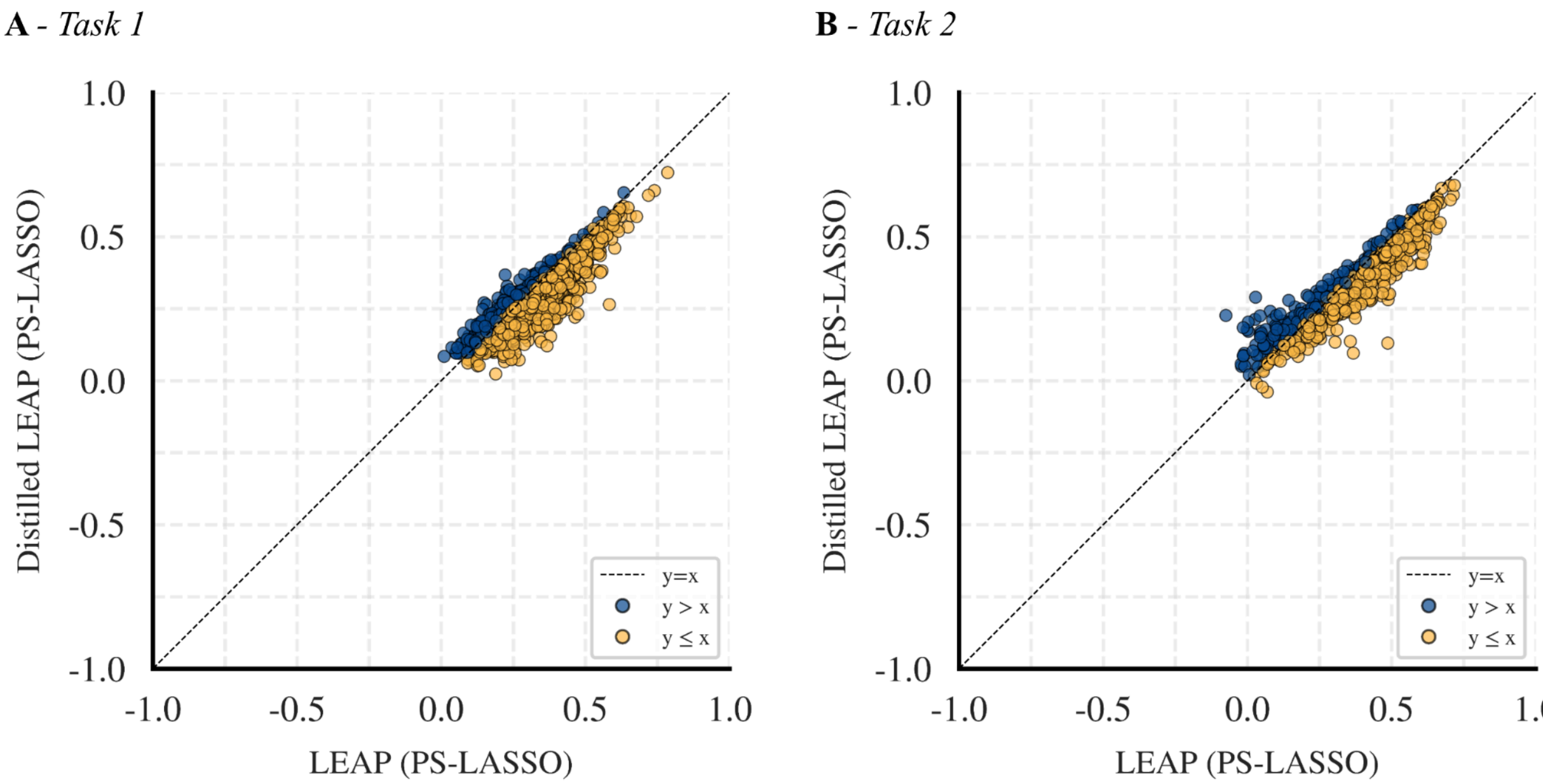

***Figure 4. Evaluation of distillation performance.*** *We compare the performances in predicting perturbation response using LEAP (PS-LASSO) (y-axis) or using PS-LASSO regressors fitted using stably selected biological pathways only as input (x-axis). This distillation is performed for tasks 1 (A) and 2 (B).*

To further evaluate this, we analyse the main Reactome pathways selected by our model for some of the top predicted compounds (Supplementary Table 9). Overall, our model successfully identified pathways with a clear and plausible link to the drug's known mechanism of action. These plausible links include pathways directly involved in the known mechanism of action of the drug or expected downstream cellular responses, based on the biological function of the target. For example, the pathway “Signalling by MAPK mutants” was successfully selected by our model for refametinib and SCH772984, which are both known to target key kinases of the MAPK signalling pathway. Similarly, analysis of the pathways selected for the PLK1 inhibitor GSK-461364 highlights a significant involvement in the ‘breakdown of the nuclear lamina’. This is highly consistent with PLK1's regulatory role in mitotic cell division, a process intrinsically linked to the precise and controlled disassembly of the nuclear lamina.In contrast, the identified pathways for triazolothiadiazine were plausible but less direct. This could reflect a wider spectrum of mechanisms for this compound, which is known to interact with various biological targets and activities. Overall, these results indicate that LEAP predictions for drug response are grounded in biologically plausible mechanisms.

# Discussion

Identifying disease models or patients who are likely to respond to a given treatment is a critical need in drug discovery. In this paper, we measure prediction performance with per-perturbation Spearman's correlation to evaluate the ability of the model to rank disease models by their response to a given perturbation. Our model, LEAP (PS-LASSO) outperforms previous approaches in unlabeled cell lines, including DeepDep (72% increase in average Spearman's correlation in task 1').

The LEAP framework consistently improves prediction performances compared to base models for all modelling approaches, with an increase in average Spearman's correlation ranging from 1.4% to 4.4%. LEAP applied to perturbation-specific LASSO regressors (PS-LASSO) yields some of the best prediction performances, with Spearman's correlations of 0.321 in task 1 and 0.350 in task 2, and provides the best balance between prediction performance and computation time. Although pan-perturbation models take additional information as input (perturbation fingerprints), they do not always outperform their perturbation-specific counterparts. We hypothesise that this could be due to (i) the use of a global loss discriminating more between perturbations than between samples, and (ii) a more challenging tuning. In particular, one pan-perturbation model requires much more parameters than one per-perturbation model, and the optimal hyper-parameter set must fit all perturbations and cannot be optimized for each perturbation.
The prediction performance obtained with LEAP (PS-LASSO) in an external validation study (average Spearman's correlation of 0.199 in PRISM) is close to the label measurement reproducibility (average Spearman's correlation of 0.220 between drug responses measured in our training studies and in PRISM), which suggests that our model captures most of the predictive signal. The observed Spearman's correlation in drug response between our training studies (GDSC, CTRP and CCLE) and PRISM remains low, despite the corrections for technical differences between studies implemented in the R package PharmacoGx. Previous publications [32,33] also commented on this low measurement reproducibility and reported Pearson's correlations between measured responses in different studies varying between 0 and 0.8 depending on the drug and studies.

We demonstrate the use of LEAP with five MAE models, each trained on all available unlabelled samples to transform the data before training the regression models. This has two main advantages. First, the use of pre-trained representation models instead of training the representation model on each training set reduces the computational time. Second, the representation model can be trained on a larger and more comprehensive RNAseq dataset. In this paper, we leverage a larger publicly available RNAseq data from cell lines and PDX models that did not necessarily have perturbation responses available. In our evaluation framework, the label (perturbation response) from the test set is unseen, but the features (RNAseq) are seen during representation model training. Given new RNAseq data where perturbation responses are unknown, our unsupervised MAE models could be retrained on all available RNAseq data, including both labelled and unlabelled samples[15]. In our ablation study, we observe a small decrease in prediction performance when using MAE models trained on all samples instead of training samples only. We hypothesize that this small decrease in performance is due to the fact that a MAE trained on all samples learns features that may be underrepresented in the training set and cannot be exploited by the regressors.

Our ablation study shows that the ensembling of regression models trained on data representations obtained with multiple MAE models in LEAP generates an increase in performance compared to the ensembling of regressors trained on a single representation. This suggests that the stochasticity of deep learning representation models can provide robustness and capture complementary patterns. Our representation learning ensemble framework may be beneficial in other machine learning tasks, including for instance predictions from histology slides.

We identify important Reactome pathways in LEAP using a novel interpretability framework based on stability selection for model distillation[11,25]. Our selected pathways are stable and yield prediction performances that are close to that of LEAP. Human review of these selected pathways suggests plausible implications based on known mechanisms of actions of the drugs. Beyond biological validation, our interpretability framework could enable a better understanding of the biological mechanisms involved in perturbation response, including the identification of alternative pathways potentially involved in resistance mechanisms[34].

Several research directions can be explored to improve on our work. First, the hyper-parameter tuning is based on grid search using restrictive grids for all approaches and some of the hyper-parameter choices could be further explored (e.g. dimensionality in MAE). This could be improved in future work by expanding the hyper-parameter search space and using a better optimized parameter sampling algorithm, provided that the computational budget is increased. Similarly, parameters that relate to the processing and representation of gene expression data could be further tuned for the specific task of perturbation response: representation dimension, normalisation, scaling, gene selection. For those parameters, we adhere with parameters selected in previous work[12]. Second, we use a simple ensembling strategy by calculating the unweighted average of predictions from multiple regressors in LEAP. Performances could potentially be improved by using a weighted average based on the performances of individual regressors, more complex ensembling techniques (e.g. routing or boosting), or a different number of ensembled models. Third, we use established perturbation fingerprints for the pan-perturbation models. Perfecting these representations may improve the performances of pan-perturbation models. Fourth, even if the hyper-parameter selection of pan-perturbation models is dictated by per-perturbation metrics, their loss and consequently their optimization rely on a global loss over all perturbations and preclinical models, which may be detrimental to their final performance. Fifth, all models presented in the present paper only use gene expression profiles as input. Although previous work suggested that gene expression is the most informative modality to predict perturbation response[10], additional data modalities, including mutations and protein expression, are available and could be leveraged to better characterize the cell lines. Sixth, there are known intrinsic differences between the molecular profiles of cell lines and PDXs. In particular, previous work suggests that the gene expression in PDX is closer to the one in tumors than the gene expression in cell lines, as cell lines suffer from the lack of cell diversity, tumor micro-environment, and sometimes high passage numbers[35]. Although we aim at exploring the generalization capabilities of LEAP by evaluating models trained on cell lines data on PDX, those known differences may partially explain the lower performances on this task.

Future work could also focus on applying LEAP to transfer learning approaches to improve prediction performances in out-of-domain data, including in held-out tissues, disease models or even patients. In particular, the LEAP framework could be applied to other models that are designed to generalize beyond the training context, including (i) TCRP[11], which uses a model-agnostic meta-learning strategy to pre-train neural networks that can be quickly adapted to new tasks with limited data (few shots), or (ii) CodeAE[15], which employs an autoencoder to align disease-model molecular data with patient data while handling potential confounding factors. Our method could also be adapted to predict binary labels[15]. This may improve robustness to noise in the labels and allow to learn common patterns across perturbation responses with different levels of granularity in cell lines, PDX and patients (e.g. from AAC to RECIST scores). If successful, we anticipate that such prediction models trained in preclinical data and adapted to a clinical setting could be instrumental in better prioritising patients and indications to include in early phases of clinical trials when patient response data is still scarce. This approach could ultimately improve the likelihood of clinical success by targeting patients who are more likely to respond well to the treatment.

# Conclusion

We propose LEAP, a new framework ensembling regressors trained over multiple deep representations of expression profiles. Our framework consistently improves performances for all prediction algorithms and tasks that we consider. In particular, LEAP (PS-LASSO) outperforms state-of-the-art models for gene essentiality and drug response prediction in cell lines. Its lower computational burden and minimal tuning requirements facilitate broader and more effective applications in the prediction of perturbation response. We demonstrate that LEAP is interpretable and captures meaningful biological information using stability selection for model distillation. We believe that our approach could accelerate the discovery, positioning and repurposing of anti-cancer therapies to ultimately improve patient outcomes. We make our code and evaluation pipeline available for future research.

# Materials and Methods

## Datasets

This study is based on publicly available data from DepMap, PharmacoDB and PDX Encyclopedia (Supplementary Table 1).

## DepMap

CRISPR knockout screens on human cancer cell lines and related omics datasets are downloaded from DepMap 23Q4. Perturbation screens' outcomes are obtained from the project Achilles[36], and processed by DepMap[5] through a pipeline [37,38] using Chronos[39] on each screen library to output gene effect scores. For every studied gene representing a dependency of interest, gene effect scores are then aggregated and corrected per cell line to generate gene dependency scores. Both scores are available on the DepMap platform. Our study focuses on the gene dependency score as it measures the probability that a gene is essential for the survival of a particular cell line. Gene expression data is obtained from CCLE[40] via the DepMap portal. We use the publicly available transcripts per million (TPM) normalized counts as cell lines descriptor inputs for our models.

For task 1' and comparison with OmicsRPZ[12], gene dependency data and TPM gene expression data are downloaded from DepMap 22Q4[41]. Perturbation outcomes and omics originate from the same sources and pipelines, without corrections added in 23Q4, but do not cover as many cell lines as in later versions.

## PharmacoDB

Drug perturbation screens on human cancer cell lines are downloaded from PharmacoDB[30] using the R package PharmacoGX (version 3.4.0)[42]. From the 15 studies that are accessible through PharmacoGX, we retain CCLE-2015, GDSC-2020 (v1-8.2 and v2-8.2), CTRPv2-2015 and PRISM-2020 which are respectively referred to as CCLE, GDSC (v1 and v2), CTRP and PRISM. The tested drugs (Supplementary Table 1) cover a wide range of anti-cancer drugs including cytotoxic agents (paclitaxel, bendamustine) and targeted therapies (erlotinib, everolimus, ruxolitinib), and range from experimental molecules to clinically-approved drugs. For the different learning tasks, we use the area above the dose-response curve (AAC) which is measured on a drug-specific dose range. A high AAC shows a high average efficacy.

Gene expression data before perturbation is downloaded with PharmacoGX when available, which is only the case for some of the cell lines studied in GDSC (v1 and v2) amongst the retained studies. For other studies or cell lines without gene expression data in GDSC (v1 and v2), we use the gene expression data from DepMap 23Q4.

## PDX Encyclopedia

Drug response screens in patient-derived xenografts (PDX) are obtained from PDX Encyclopedia[31]. Of the 63 available perturbations, we restrict our analyses to the five mono-therapy drugs that were studied previously[11]: Paclitaxel, Cetuximab, Erlotinib, Tamoxifen and Trametinib. Drug response is provided as the minimal percentage of tumor volume change observed over 10 days of treatment. We download available gene expression raw counts.

# Perturbation fingerprints

In the pan-perturbation model approach, perturbations' encodings are required as a unique model learns and predicts the outcome of diverse perturbations. In practice, such models use gene and drug biological or molecular descriptors (also called fingerprints) to better capture similarities between perturbations[44].

## Gene fingerprints

We use gene fingerprints derived from the Molecular Signatures Database (MSigDBv6.2[45]) that describe the affiliation of each gene (dependency of interest) in gene sets that are up or down-regulated in response to a perturbation experiment. Most of those signatures are curated from the literature. Gene perturbations with similar fingerprints tend to be activated or repressed together in perturbation experiments included in the Molecular Signatures Database. We use the same 3115 chemical and genetic perturbations (CGP) curated gene sets as in DeepDEP[10], which were obtained after filtering out the 318 sets that do not involve any of the 1298 genes studied in their paper.

## Drug fingerprints

Drug fingerprints are solely based on their molecular composition. Every drug perturbation name obtained from PharmacoDB[30] or PDX Encyclopedia[31] is mapped to its corresponding Compound Identifier (CID) on PubChem via the PubChem API[46]. Canonical Simplified Molecular Input Line Entry System (SMILES) are then downloaded from the compound card and further processed with the github repository CDDD [47] and the rdkit package ("RDKit: Open-source cheminformatics") to remove salts and stereochemistry from the SMILES sequence. Additionally, we keep existing filters in the CDDD package that filter out molecules based on their molecular weight, number of heavy atoms or partition coefficient. Morgan fingerprints are then generated from SMILES with rdkit as vectors, using a radius of 2 and a size of 2048.

As the SMILES are downloaded from PubChem, perturbations that are combinations of compounds or drugs that do not match to any CID (lack of matching terms, inexistence of a compound card for compounds in early development stages) cannot be encoded with fingerprints and are excluded from tasks that compare the performances of pan-perturbation versus perturbation-specific model approaches.

# Data preparation

## Sample and perturbation selection

Overall, the data curation of DepMap[5], PharmacoDB[30] and PDX Encyclopedia[31] is done in two steps where (i) we keep samples with available RNAseq data, information on the tissue of origin and exclude outliers, (ii) we select the perturbations that will be investigated (Supplementary Figure 1). For PharmacoDB, we have an additional final step to remove duplicated sample-perturbation pairs arising when combining data from GDSC, CCLE and CTRP.

**DepMap**. For our main tasks, we use the version of DepMap released in Q4 of 2023 (version 23Q4). We drop 79 DepMap cell lines with no available RNAseq data (Supplementary Figure 1A). Two outliers are then identified in the DepMap data by visually inspecting the first principal components obtained from a Principal Component Analysis (PCA) of the RNAseq data. After excluding these outliers, we define a new list of dependencies of interest with standard deviation of the gene dependency above 0.2[10], resulting in 1,593 dependencies. For comparison with previous studies, we also reproduce our analyses using a former version of the DepMap data in task 1' using 893 cell lines and 1,223 dependencies of interest that were used in OmicsRPZ[12].

**PharmacoDB**. We first prepare the data from each of the five retained studies individually (Supplementary Figure 1B). We only keep the drugs with available fingerprints to allow for the comparison of different modeling approaches (see below). We then remove duplicated responses measured for the same drug in the same cell line across the five studies. More precisely, within each single study, when the same drug is tested multiple times on the same cell line, we aggregate the observations by taking their median response. This operation helps reduce uncertainty and noise in experiments performed in the same study. Later, if the same drug is tested on the same cell line in different studies, observed outcomes are not aggregated across those studies as experimental setups such as concentration ranges may vary. Therefore, we keep only one of the studies' outcomes by taking the first appearance of the experiment based on an arbitrary order of the studies: GDSC v2, GDSC v1, CCLE, CTRP and PRISM. Unless a precise study is explicitly mentioned, we refer to the aggregation of the 4 datasets GDSC v1, GDSC v2, CCLE and CTRP as PharmacoDB. For all learning tasks, we only retain drugs that are administered to at least 75 cell lines. We use PRISM as an external validation set and retain drugs that are administered to at least 15 cell lines in this dataset.

**PDX Encyclopedia**. We keep the 140 PDX with available RNAseq and drug response data (Supplementary Figure 1C). Of the 63 treatments tested in the study (including 25 combinations of drugs), we only keep the 5 drugs that were also administered to cell lines in the PharmacoDB dataset [11].

## RNAseq data preprocessing

We download TPM counts from DepMap[5] and transformed raw counts into TPM normalized counts using an in-house software package for GDSC[6] and PDX Encyclopedia[31]. TPM counts are then log2-transformed with the addition of a pseudo count of 1. We keep the 16,926 genes available in all three studies and concatenated the samples from the three RNAseq datasets. To correct for study effects while preserving biological differences due to the tissue of origin, we apply combat[49] as implemented in InMoose[50] to log2-transformed TPM counts. We then filter the 5,000 most variant genes. We standardize the TPM counts, so that each gene expression has a mean of zero and a standard deviation of one.

## Tasks

We define three types of challenges, where we aim to predict perturbation response in (i) held-out cell lines, (ii) an held-out tissue, or (iii) an held-out disease model (Table 1)[11].

### Challenge I: prediction in unscreened cell lines

The first challenge encompasses two tasks that focus on predicting the response of a new cell line to a perturbation (a gene knockout or a drug) already performed on other cell lines. From a clinical perspective, this challenge mimics a key approach in precision oncology: to determine if a new patient would respond to a therapy that was already tested in other patients.

The first task consists in predicting gene dependency in cell lines using the DepMap[5] dataset. For this, we train and evaluate models to predict the dependency of each of the 1,539 genes defined above using the 23Q4 release of the DepMap dataset that was curated as described above (task 1). For an exact comparison with OmicsRPZ[12], we also implement an alternative version of this task using the same version of the DepMap dataset that was used in OmicsRPZ (task 1') where we aim to predict the dependency of 1,223 genes.

The second task focuses on the prediction of drug response (AAC) in cell lines using the PharmacoDB[30] dataset (task 2). Following recent recommendations[51], we focus on AAC instead of other measures like IC50. The models are trained and evaluated in GDSC v2 and v1, CTRP and CCLE, and we keep PRISM as an external test set. We predict the response to the 648 drugs available in the training set (GDSC, CTRP and CCLE) that are administered to at least 75 cell lines (Supplementary Figure 1). The evaluation of performances in PRISM is restricted to the 254 drugs that are also included in the training data.

To evaluate the prediction performance of the different modeling approaches, we use a repeated holdout cross-validation framework [12]. The data is split into a training set with 80% of the cell lines and a test set with the 20% remaining cell lines. This procedure is repeated 10 times to increase the robustness of the evaluation.

### Challenge II: prediction in an unscreened tissue

The second challenge comprises two tasks that aim at inferring the response to perturbation (a gene knockout or a drug) in cell lines originating from a tissue that is not observed in the training set (Table 1). This challenge could be seen as a drug repurposing task, where we aim to evaluate if an existing drug might be effective for a new indication or a different subpopulation of patients. It is also valuable for target identification and validation, especially when genes or drugs have been screened across various cell lines, but not specifically in the tissue of interest. For instance, in DepMap[5], only 24 cell lines are available as a proxy for mesothelioma, and only 3 of sarcomatoid type. These out-of-domain tasks are more arduous as models aim to predict in cell lines with tissue-related molecular specificities that are not seen at training time.

For the gene dependency task 3, the out-of-domain capabilities of the different models are assessed on each of the 18 tissues with more than 25 cell lines available in the latest DepMap data, so that there is always a minimum of 10 cell lines available for few-shot learning, and 15 cell lines for computing a performance metric.

For drug response, we first keep the 10 tissues with the largest sample size in the PharmacoDB[30] data (task 4). The tissue selection slightly differs from task 3 as most drugs are not tested in all cell lines, which requires us to set a threshold on both the number of tissues and the number of cell lines tested for each drug within a tissue to select test cases. As a result, we use the 525 drugs that are tested in at least 25 cell lines for each of these 10 tissues (see supplemental material).

We evaluate prediction performance in cell lines from the tissue of origin that was not used in the training set. To evaluate variability, we create 100 test sets by repeatedly removing 10 cell lines from the target tissue at random. We repeat this procedure such that each tissue is in turn considered as the target tissue. We report the tissue-specific performances (average of the per-perturbation performance on a given tissue over all perturbations and test sets) as well as the average performance over all held-out tissues.

### Challenge III: prediction in an unscreened disease model

Challenge 3 aims to evaluate the ability of the models trained in cell lines to generalize to other disease models (Table 1). For this, we first train the models to predict drug response using all available cell lines in the full PharmacoDB[30] dataset (task 5). These models are then used to predict the negative minimum change in tumor volume compared to the pre-treatment baseline over a 10-day post-treatment period measured in PDX treated with the same drug.

To assess variability and be as comparable as possible to published results[11], we evaluate prediction performance over 100 randomly sampled subsets of the PDX data obtained by removing 10 samples from the PDX data (the ones used in training in a few-shot setting) and keeping the remaining ones as a test set. Of note, the number of test samples depends on the perturbation, as all treatments are not measured in all PDX.

## Machine learning framework

We aim to predict perturbation response (gene dependency or drug response) based on gene expression data. Some of our models additionally use perturbation-specific fingerprints as input.

### Overall modeling approach

To predict perturbation response, we use a regression algorithm which takes a lower dimensional representation of the data as input. We compare two types of regression approaches: a perturbation-specific approach based on multiple models (PS) where independent regressors are trained to predict the response for each perturbation, and a pan-perturbation model approach (PP) where we predict the response to any (perturbation, disease model) pair based on the gene expression of the disease model and the fingerprints of the perturbation. Our approach can be decomposed into three steps where we (i) train a Masked Auto Encoder (MAE) to represent the RNAseq data, (ii) extract the first 500 components from a Principal Component Analysis (PCA) of the fingerprint data (for the pan-perturbation approach only), and (iii) train a regression algorithm using the data embeddings as input[12]. Under the perturbation-specific (PS) paradigm, we consider baseline K-nearest neighbors (PS-KNN), LASSO regressors (PS-LASSO), and LightGBM models (PS-LGBM), each trained on a single perturbation's data. We also include three pan-perturbation (PP) methods: PP-LGBM and tDNN, with the latter further adapted into PP-MLP. For a fair comparison of the regression approaches, all methods are integrated with our MAE in place of their original autoencoders. To improve the prediction performance of all models (except for the baseline PS-KNN), we use ensembling by aggregating the predictions obtained with multiple regressors trained on different subsets of the data. In LEAP, we further ensemble multiple regressors trained on different data representations obtained with different initializations (Figure 1).

## Gene expression representation algorithm

We train representation learning models on RNAseq data to generate embeddings for the disease models (cell lines or PDX). These representation models are agnostic to the perturbation as they only depend on the molecular profiles of the disease models. The OmicsRPZ[12] study demonstrated superior performance in downstream prediction tasks by employing a Masked Auto-Encoder (MAE). We use the same MAE with hyper-parameters recommended by the OmicsRPZ study for RNAseq representation.

Briefly, we start from the method developed in VIME[52] and we adapt it to (i) remove the mask-predict pretext-task, which didn't improve the prediction performance of the downstream task but also significantly increased training time (results not shown), and (ii) optimize the mask allocation code. For each training batch, we randomly mask 30% of the input entries in the batch data matrix and use the VIME corruption method to permute these masked entries with values coming from other samples. We fix the representation dimensionality and several other parameters of the MAE based on the settings used in the OmicsRPZ paper. The representation dimension is fixed at 256, with 512 units in the first hidden layer, and training capped at 1000 epochs with early stopping (patience=20, delta=10**-5).

Since we are in a transductive task setup, and to reduce computational burden, we do not train distinct MAE models on the different training sets used for each of the 5 tasks. Instead, we train the MAE models once using all available unlabelled samples with RNAseq measurements across the DepMap[5], PharmacoDB[30] and PDX Encyclopedia[31] datasets, totalling 1920 cell lines and 191 PDX. The five MAE models trained on the same data but with different initializations are then used for all tasks (Figure 1).

In LEAP, we propose to use an ensemble of 5 MAE models obtained with different random seeds to leverage variability in this procedure and eventually increase robustness and prediction performance[20]. The change in random seed affects the selection of the early stopping holdout set, masking, and augmentation operations, and the initialization of the different layers. The 5 MAE models are trained on the same dataset including all unlabelled samples.

## Regression algorithms

We consider seven modeling approaches. For a fair comparison across the prediction models approach, we use them jointly with our MAE and consider limited grids of up to 10 combinations of parameters. In methods that incorporate already an AE, we replace their AE with our MAE.

The perturbation-specific K nearest neighbor[53] (**PS-KNN**) constitutes a baseline in our study. In PS-KNN, the response to a given perturbation is predicted as the average response observed for the K nearest cell lines based on their gene expression profile which have available response to the perturbation of interest. We consider the K=5 nearest neighbors.

In **PS-LASSO**, we train one LASSO[54] model for each perturbation. Each of the regressors in PS-LASSO takes molecular features (RNAseq) of the cell lines as input. The hyper-parameters of each of the regressors in PS-LASSO are tuned independently by maximizing prediction performance for the corresponding perturbation. We consider a small grid of 10 automatically defined values of the L1-regularization parameter using the sklearn library.

Similarly, we train one LGBM for each perturbation in **PS-LGBM**. This LGBM has 400 estimators with a maximum depth of 10 and 31 leaves. We tune the learning rate and the proportion of features to keep in each subsample. We consider a limited grid of two values for the learning rate (0.005 and 0.01) and five values for the feature proportion (0.05, 0.1, 0.15, 0.2 or 0.25).

We also train one MLP for each perturbation in **PS-MLP**. This MLP is trained with a surrogate loss for optimizing the Pearson's correlation between predicted and observed perturbation responses. We use a dropout rate of 0.2, a maximum of 200 epochs with early stopping and a learning rate scheduler. We tune (i) the initialisation of the learning rate using a grid of five values (0.0005, 0.001, 0.002, 0.005, 0.01), and (ii) the number of hidden layers, considering either one or two hidden layers of size 20.

In **PP-LGBM**, we use a pan-perturbation regression model to predict the response to any perturbation in a given cell line based on both sample-level features (RNAseq) and perturbation-level features (fingerprints) [10,12]. As in the OmicsRPZ study [12], we use an LGBM with 500 estimators, a maximum depth of 20 and 4,000 leaves, and we tune the learning rate and the L1-regularization parameter. We considere a limited grid of five values for the learning rate (0.01, 0.02, 0.03, 0.04 and 0.05) and two values for the regularization parameter (0 or 1).

We also incorporate **PP-tDNN**[18], a pan-perturbation deep neural network model designed to integrate gene expression and drug fingerprints through two subnetworks. Each subnetwork has three dense hidden layers, whose outputs are concatenated and passed through four additional hidden layers with consecutive halvings of nodes. Dropout layers follow all but the last hidden layer. Initial parameters and architecture are as defined in previous work[18]. We use a grid of five values for the dropout rate (0, 0.1, 0.25, 0.45, 0.7).

Finally, we introduce **PP-MLP**, a version of tDNN tuned to enhance performance in our study. Unlike the original tDNN, PP-MLP employs tailored hyperparameter settings and additional regularization to improve predictive accuracy across perturbations. In particular, we use hidden layers of sizes (512, 256, 128, 64, 32, 16). As in PS-MLP, we use a dropout rate of 0.2, a maximum of 200 epochs with early stopping and a learning rate scheduler. We further use batches, and tune the number of batches to be either 2048 or 8192. We also tune the initialisation of the learning rate considering the same grid of five values (0.0005, 0.001, 0.002, 0.005, 0.01) and use the same loss as in PS-MLP.

### Hyper-parameter tuning procedure

We use grid search to tune the hyper-parameters of all regression models, except the baseline PS-KNN. For each combination of hyper-parameters, we perform a 5-fold cross-validation within the training data, maximizing the per-perturbation Spearman's correlation.

For PS approaches, we tune the hyper-parameters of each perturbation-specific regression model separately. For PP approaches, we tune the hyper-parameters of the pan-perturbation regression model by maximizing the average per-perturbation Spearman's correlation. In all cases, we use 5 folds made of non-overlapping sets of cell lines.

### Layered ensembling

To improve prediction performances in LEAP, we use an ensemble model where the final prediction is defined as the average of the predictions obtained from multiple regressors. For this, we first use the 5 MAE models to generate 5 different representations of the training data. The regressors trained in each iteration of the cross-validation conducted on each of the 5 representations of the data are used for ensembling. A total of 25 regressors are ensembled, coming from the data embeddings of 5 MAE models and for each, 5 regressors trained in a 5-fold CV.

## Performance metrics

Prediction performance is measured by comparing the true and predicted perturbation response in the test set of unscreened cell lines (challenge I), cell lines from an unscreened tissue (challenge II), or unscreened PDX (challenge III).

As in previous work, we use the Spearman's and Pearson's correlations as performance metrics [12]. We report both (i) the average of per-perturbation correlations measured by comparing the true and predicted responses to each of the perturbations, and (ii) the overall correlation measured by comparing the true and predicted response for all sample-perturbation pairs in the test set. Although overall metrics are not appropriate in our setting as they primarily capture differences between the perturbations rather than differences between samples[55], we report them in Supplementary Materials for comparison with previous studies. As small variations in ranking may not always be meaningful, we also report the per-perturbation Mean Squared Error (MSE) as a complementary metric.

# Biological interpretability

## Interpretability framework

To generate biologically interpretable explanations for LEAP predictions, we transform gene expression profiles into pathway-level representations. Specifically, for each Reactome pathway, we perform principal component analysis (PCA) on the gene expression matrix restricted to that pathway's genes, and use the first principal component (PC1) as a continuous pathway expression feature[24].

We then distill LEAP by fitting perturbation-specific LASSO models regressing predicted values from LEAP as a function of pathway-level features. To assess the robustness and consistency of pathway selection, we apply stability selection [25,56,57]. Briefly, LASSO regressors are fitted on 100 subsamples of the training data to compute selection proportion for each of the pathway features. Pathways that are consistently selected across subsamples are considered stable predictors. The LASSO regularisation parameter and threshold in selection proportion are calibrated by maximising the sharp score measuring model stability[56]. We report a normalised sharp score, where a value of 1 would indicate that the same features are selected in all regressors fitted on 100 different subsamples of the training data.

## Distillation validation

Finally, to validate the explanatory power of the selected pathways, we train reduced LASSO models using only the most stable pathways as input. We compare their predictive performance to LEAP models trained for the same perturbation. High correlations between reduced and full model performance would indicate that selected pathways capture a substantial portion of the predictive signal driving LEAP's predictions.

# Contributions

- Conceptualization and Research Direction: AD, BB, GD, JK, ED
- Methodology and Evaluation Design: AD, BB, GD
- Software Development and Review: BB, GD, AD, LTM
- Formal Analysis – Model Development (Representation and Ensembling): AD, GD
- Formal Analysis – Model Development (Prediction Models, Tuning): BB, GD, AD
- Formal Analysis – Model Development (Interpretability): BB, LTM
- Human Evaluation (Interpretability): RC
- Investigation and Experiments: BB, GD, AD
- Visualizations: GD, BB, LB
- Writing: BB, AD, GD, LTM, LB
- Supervision and Coordination: AD

# Declaration of Competing Interest

All authors are employees of Owkin, Inc., New York, NY, USA.

# Data availability

All the datasets used in this paper are publicly available.

# Code availability

The code used in this paper is publicly available at https://github.com/owkin/leap.

# Acknowledgements

We want to thank Jean-Philippe Vert, Floriane Montanari, Mathieu Andreux and Geneviève Robin for their scientific guidance and feedback on the manuscript. We thank Elodie Pronier and Rita Santos for their recommendations on the choice of experimental labels. We thank Nicolas Dop for reviewing the implementation of the distillation validation, Chiara Regniez for discussions on data alignment, and Maria Telenczuk for optimising the memory usage of pan-perturbation models. We are also grateful to contributors involved in earlier phases of the project: Baptiste Gross provided guidance on the use of representation models, Jean El Khoury contributed to the initial ensembling tests, Khalil Ouardini implemented the initial masking, Quentin Klopfenstein and Hugo Malafosse assisted with testing and tuning initial perturbation-specific models.

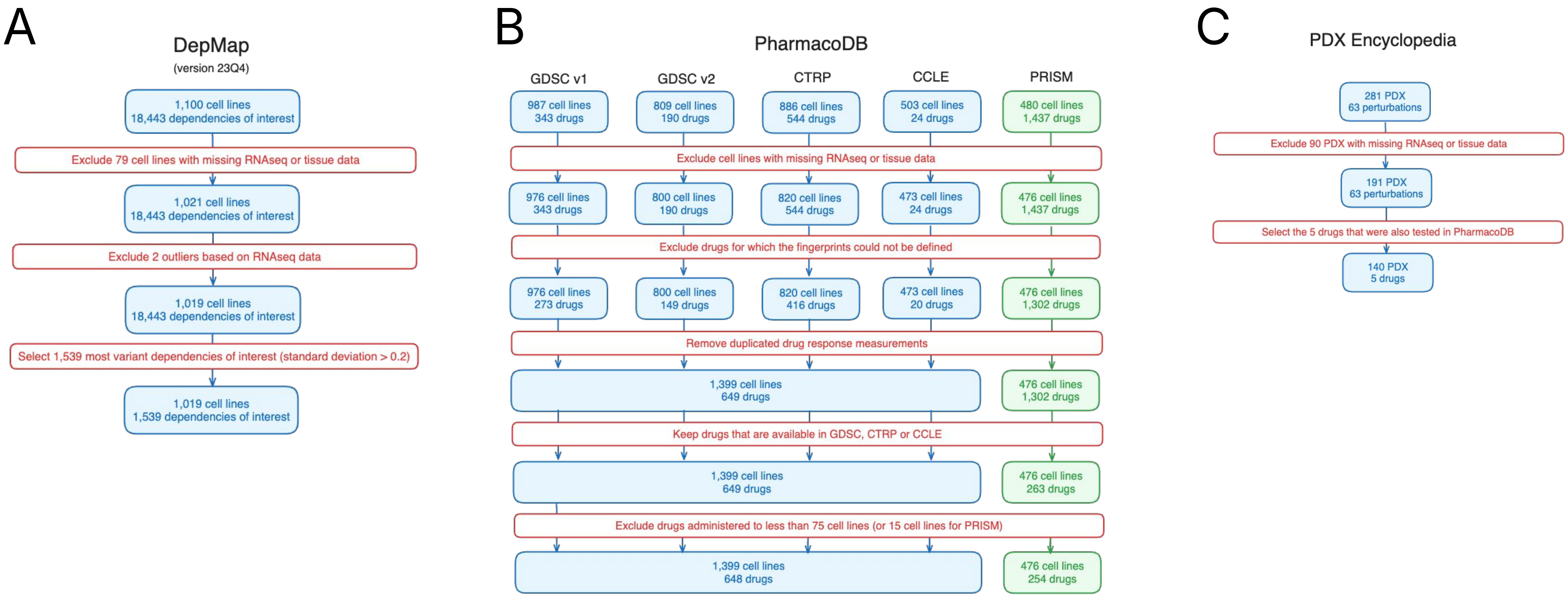

**Supplementary Figure 1. Flowcharts illustrating the data preparation steps.** We report each step of our procedure to select cell lines and perturbations in DepMap (A), PharmacoDB (B) and PDX Encyclopedia (C).

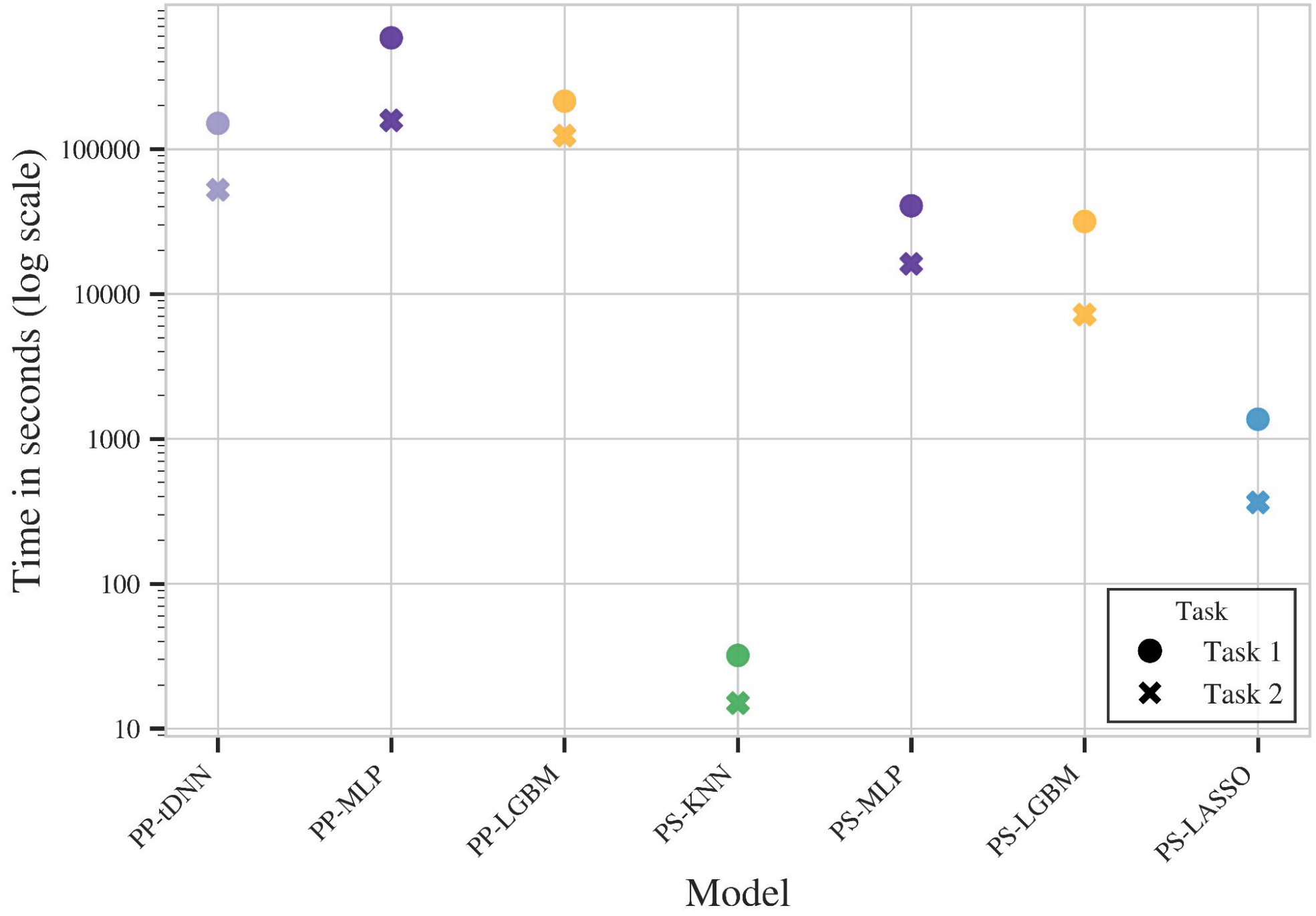

**Supplementary Figure 2. Computation times for the different models on tasks 1 and 2.** The reported computation times are for the training and testing in the 10 holdout test sets repeats. All LEAP models require 5 more computation time as they are trained using 5 different MAE. Models trained on GPUs (PP-tDNN, PP-MLP and PS-MLP) are trained on the AWS instance ml.g4dn.12xlarge, models trained on CPUs are trained on the AWS instance ml.m6i.32xlarge.

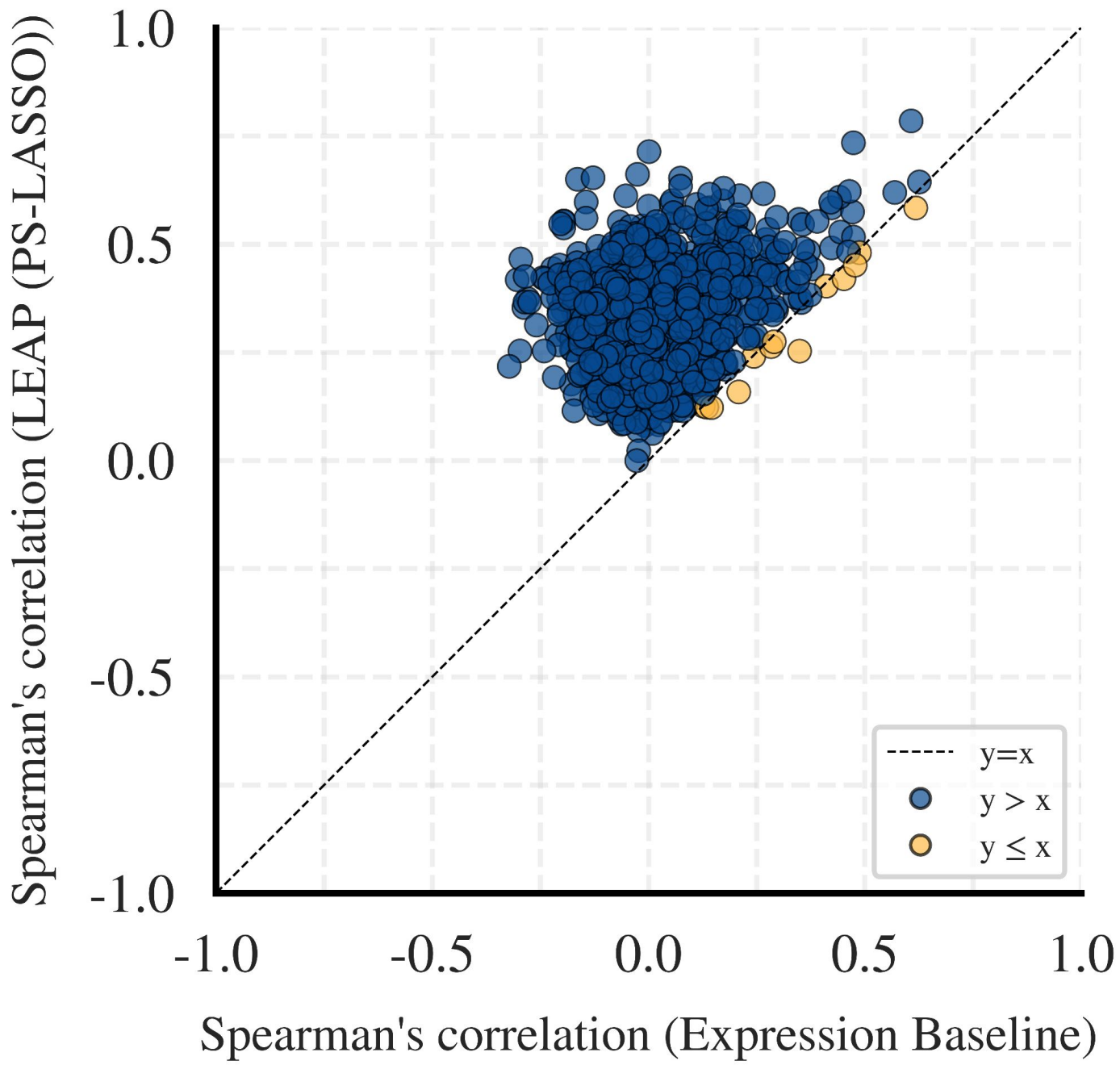

**Supplementary Figure 3. Comparison of prediction performances using LEAP against a baseline using target expression only.** On the task 1, for each of the 1,539 gene perturbations in DepMap, we considered a baseline approach where the cell lines are ranked based on the expression of the gene that is knocked-out. We compare the average per-perturbation Spearman’s correlation over the 10 training-test splits using this baseline (x-axis) or LEAP (y-axis). Of the 1,539 gene perturbations, only 13 can be better predicted using the expression of the perturbed gene in the cell line.

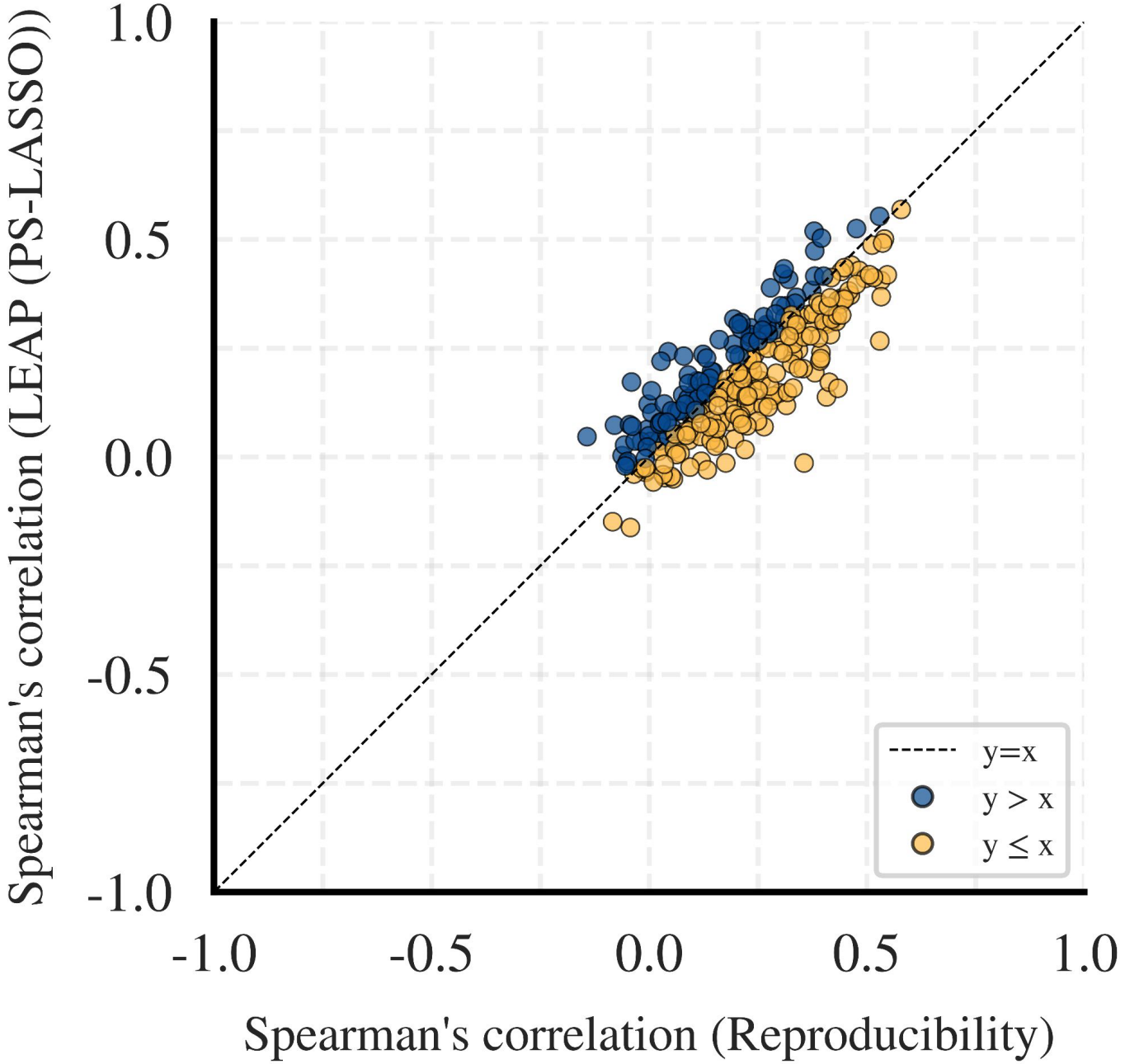

**Supplementary Figure 4. Comparison of prediction performances of LEAP (PS-LASSO) with drug response reproducibility across studies.** For each of the 257 drugs, label reproducibility is quantified by the Spearman's correlation in the AAC response to the drug measured in the same cell lines in train/test studies (GDSC, CTRP and CCLE), and the external validation study (PRISM). To ensure that models are tested on unseen cell line-drug pairs, we exclude the training cell lines from the external validation study (PRISM). This procedure is repeated 10 times and the average performance for the drug over those repeats is shown here. Of the 257 drug perturbations, 96 can be better predicted in PRISM using LEAP (PS-LASSO) rather than looking at the reproducibility from other studies.

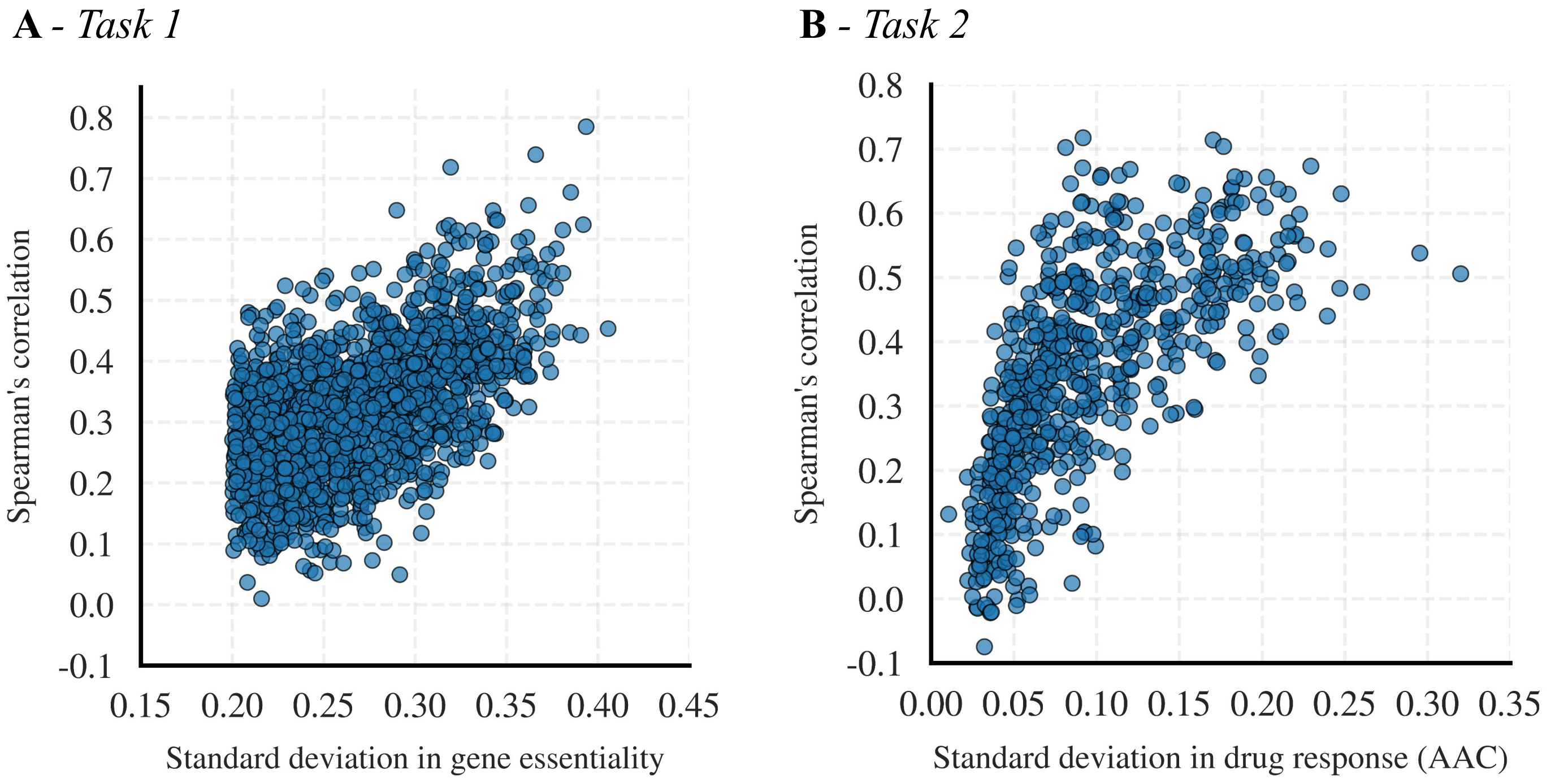

**Supplementary Figure 5. Comparison of perturbation-specific performance and standard deviation in perturbation response.** We compare perturbation-specific prediction performances measured by the Spearman's correlation in task 1 (A) and 2 (B) with the standard deviations in gene essentiality (A) and drug response (B), respectively.

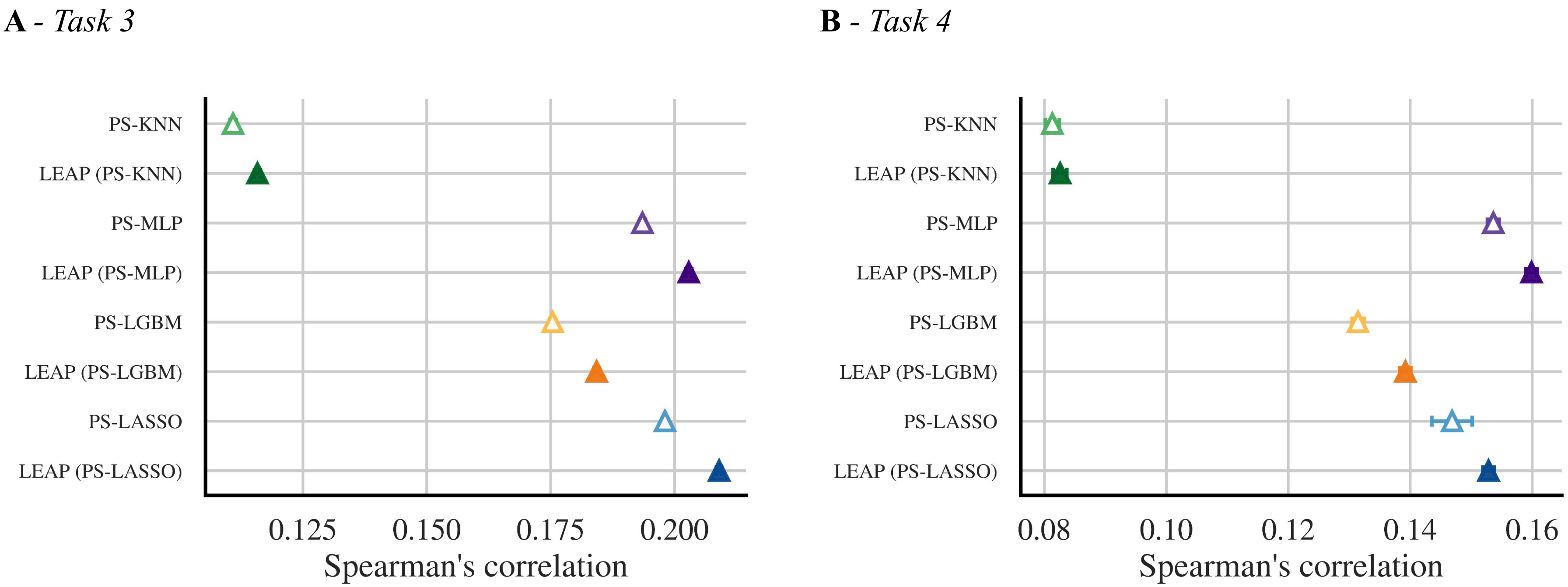

**Supplementary Figure 6. Prediction performances obtain with perturbation-specific (PS-) models on out-of-domain tasks 3 and 4.** The average spearman correlation over all perturbations and all tissue test sets is reported for task 3 (A) and task 4 (B). The error bars correspond to the standard deviation of the performance observed when sampling test cell lines in 100 different ways, removing 10 cell lines at a time per tissue.

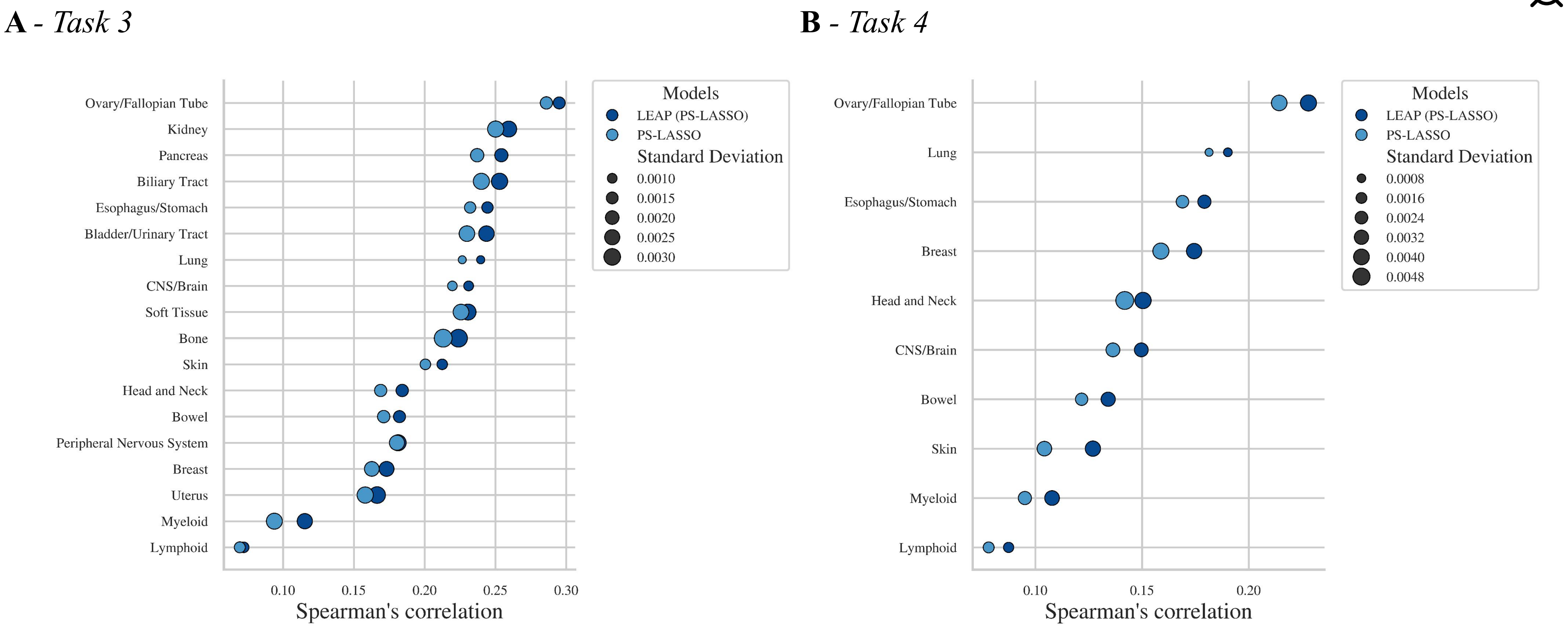

**Supplementary Figure 7. Performance in predicting perturbation response by tissue for PS-LASSO and LEAP (PS-LASSO)**. We evaluate the performances in predicting gene dependency (A, task 3) or drug response (B, task 4) in cell lines from an unseen tissue. For each tissue, we report the mean Spearman's correlation between observed and predicted responses per perturbation, averaged over 100 test sets where 10 cell lines of the tissue are randomly removed. The size of each data point is proportional to the standard deviation on those 100 test sets.

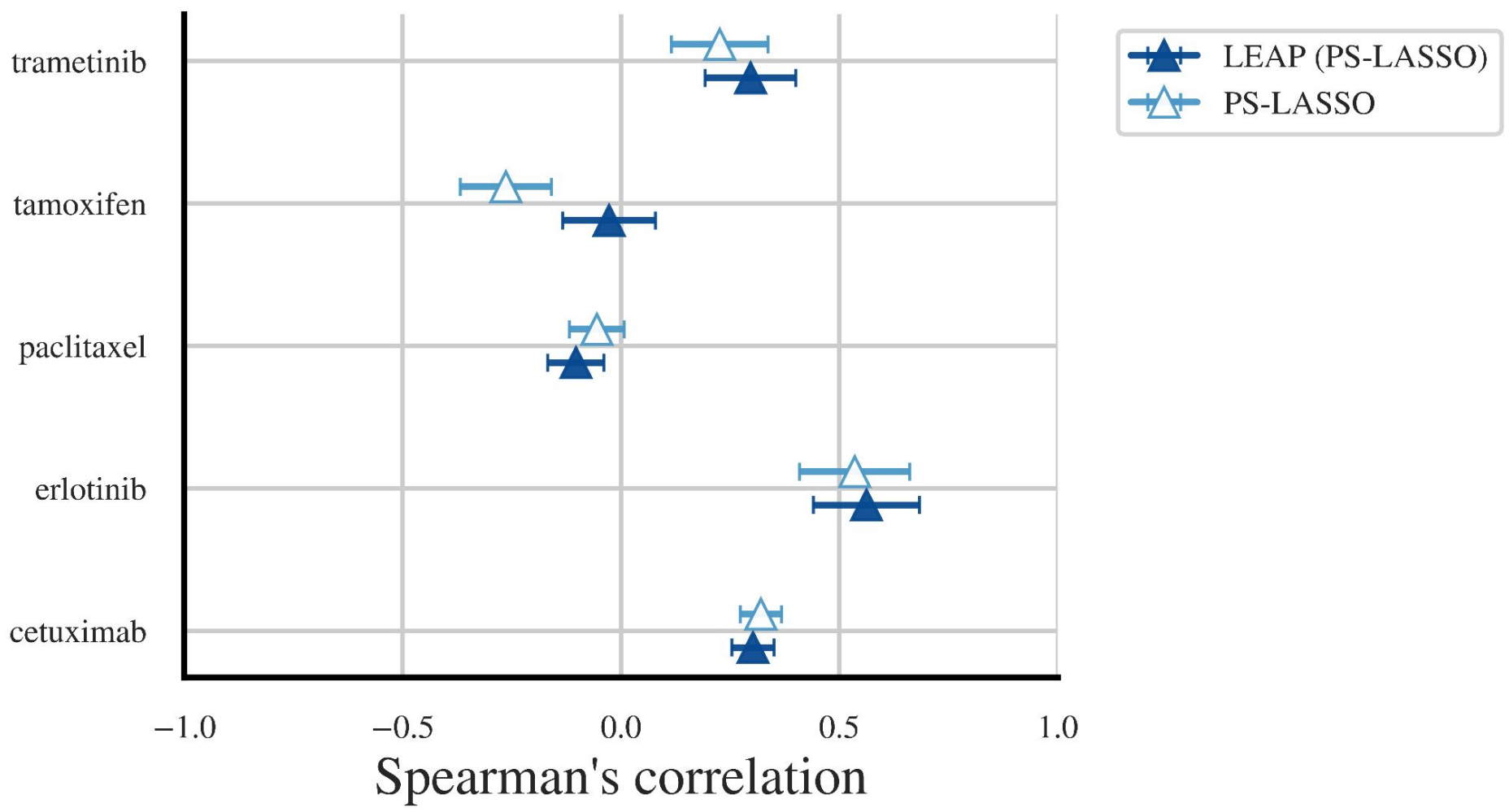

**Supplementary Figure 8. Performances in predicting the response to 5 drugs in PDX samples (task 5) with PS-LASSO and LEAP (PS-LASSO).** To assess variability, we report the mean (standard deviation) of these metrics calculated over 100 subsets of the data obtained by removing 10 PDX samples.

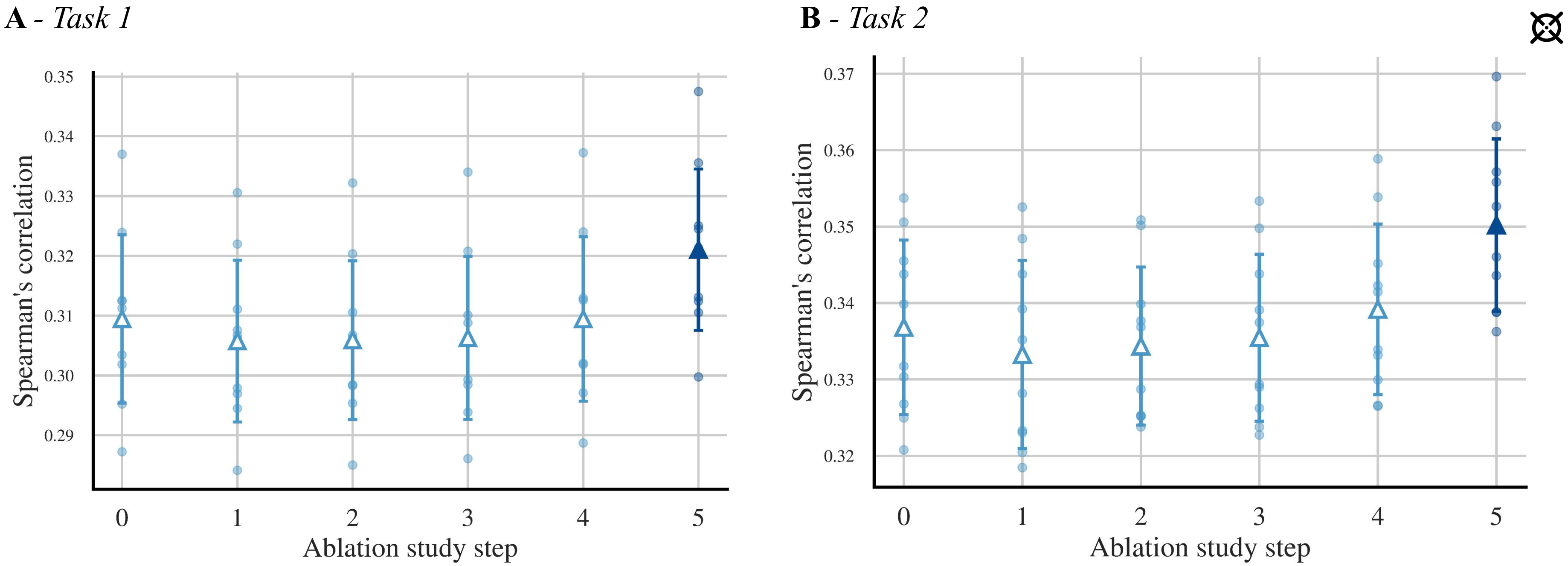

**Supplementary Figure 9. Change in prediction performances in tasks 1 and 2 due to each of the methodological choices.** Performances are reported on 10 different held out test sets and represent the average performance per perturbation. In step 0, we use the PS-LASSO model with a unique preprocessor and MAE representation model trained on each of the 10 training sets. In step 1, we pre-train a single preprocessor and MAE model on all DepMap transcriptomics data that are later used to obtain the cell line representations in all trainings. In Step 2, the preprocessor and MAE are pretrained on a concatenation of transcriptomics data from DepMap and GDSC. In Step 3, transcriptomics from PDXE are added to the pretraining data. In Step 4, 5 PS-LASSO prediction heads trained in 5-fold cross-validation splits of the transformed gene expression representation are ensembled. Finally, in step 5, we repeat the previous ensemble training of the 5 PS-LASSO models using 5 different MAE (where initialisation varies) representations, leading to the LEAP (PS-LASSO) model. The results are shown on the task 1 (A) and on the task 2 (B).

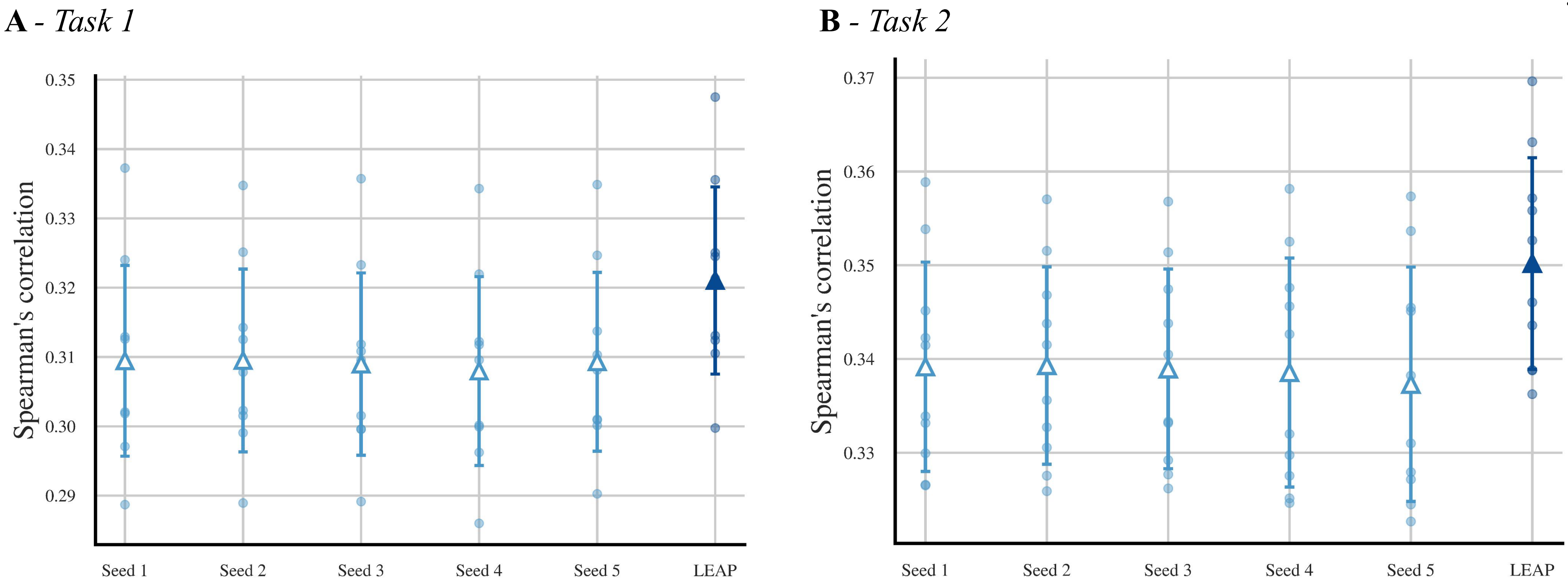

**Supplementary Figure 10. Prediction performances obtained with PS-LASSO using each of the 5 seeds or with LEAP.** We report the performances obtained with each of the 5 PS-LASSO models using different DAMAE models. Prediction performance is measured by the Spearman's correlation between observed and predicted gene dependencies in task 1 (A) and in task 2 (B).

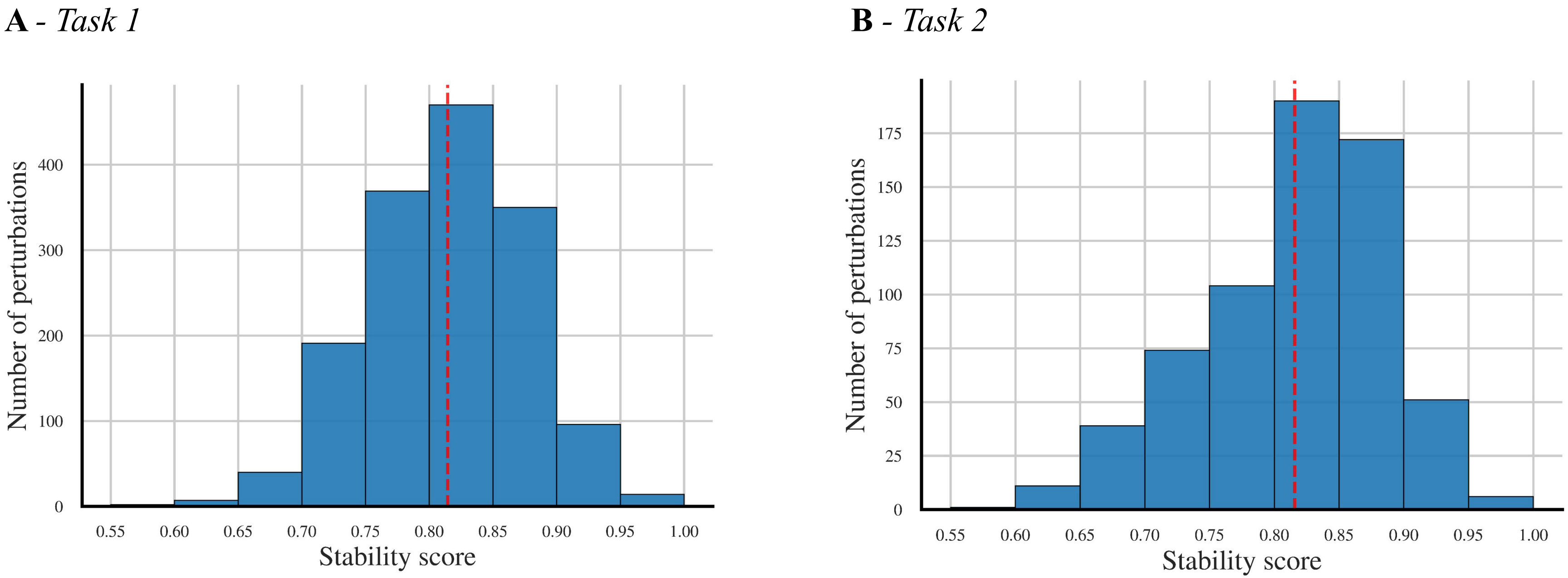

**Supplementary Figure 11. Histograms of stability scores from distillation models using biological pathways.** Stability selection is used to identify important biological pathways in LEAP (PS-LASSO) to predict each of the 1,539 genes in task 1 (A) and 648 compounds in task 2 (B). We visualise the distribution of the best normalised stability score (x-axis) obtained from each of these stability selection algorithms.

**Supplementary Table 1. Overview of the available data for model training and evaluation (after data preparation).**

| Study | Sub-study | Disease model | Label | RNAseq data acquired | Number of tissues | Number of samples | Number of perturbations | Number of sample x perturbation pairs | Number of dose points (drug response only) | Overall fold range (drug response only) |
|---|---|---|---|---|---|---|---|---|---|---|
| DepMap | 23Q4 | Cell line | Gene dependency | Yes | 27 | 1,019 | 1,539 | 1,568,222 | - | - |
| | 22Q4 | | Gene dependency | Yes | 27 | 893 | 1,223 | 1,092,139 | - | - |
| | 18 (first release) | | Gene effect | from 23Q4 | 20 | 335 | 469 | 156.646 | - | - |
| PharmacoDB | GDSC v1 | Cell line | Area Above the drug dose-response Curve (AAC) | Yes | 28 | 976 | 343 | 284,823 | 9 or 5 | 256 |
| | GDSC v2 | | | Yes | 26 | 800 | 176 | 127,053 | 7 | 1,024 |
| | CTRP | | | No | 25 | 820 | 544 | 346,032 | 16 | 32,768 |
| | CCLE | | | No | 23 | 473 | 24 | 10,757 | N/A | N/A |
| | PRISM | | | No | 21 | 476 | 254 | 644,437 | 8 | 2,187 |
| PDX Encyclopedia | | Patient-derived xenograft | | Yes | 4 | 140 | 5 | 227 | - | - |

**Supplementary Table 2. Performances in predicting perturbation response in unseen cell lines and unseen tissue.** We compare the performance of our novel approach (LEAP) with all base models. We evaluate the performances in predicting gene dependency in unseen cell lines (task 1), drug response in unseen cell lines (task 2), drug response in unseen cell lines from the external validation dataset (task 2 PRISM), gene dependency in an unseen tissue (task 3), and drug response in an unseen tissue (task 4). For each data split, we calculate the average Spearman's correlations, Pearson's correlations and Mean Squared Error comparing observed and predicted responses per perturbation. We report the mean (standard deviation) of these metrics calculated over the training-test splits. We also report the Mean Rank of the different performances over the 10 splits. A Mean Rank of 1.0 indicates that the model outperforms all other models on each of the 10 splits for the given metric.

| | | Per-perturbation | | | | | | Overall | | | | | |
|---|---|---|---|---|---|---|---|---|---|---|---|---|---|
| | | Spearman's correlation | | Pearson's correlation | | Mean Squared Error | | Spearman's correlation | | Pearson's correlation | | Mean Squared Error | |
| Task | Model | Mean (Std) | Mean Rank | Mean (Std) | Mean Rank | Mean (Std) | Mean Rank | Mean (Std) | Mean Rank | Mean (Std) | Mean Rank | Mean (Std) | Mean Rank |
| | PP-tDNN | 0.2797 (0.0148) | 12.0 | 0.2932 (0.0167) | 12.0 | 0.0673 (0.0010) | 8.0 | 0.6470 (0.0070) | 10.0 | 0.6484 (0.0067) | 10.0 | 0.0673 (0.0010) | 8.0 |
| | LEAP (PP-tDNN) | 0.2920 (0.0150) | 10.8 | 0.3055 (0.0167) | 10.8 | 0.0664 (0.0009) | 7.0 | 0.6527 (0.0068) | 7.7 | 0.6539 (0.0065) | 7.8 | 0.0664 (0.0009) | 7.0 |
| | PP-MLP | 0.2973 (0.0125) | 10.0 | 0.3103 (0.0143) | 10.2 | 6.0326 (6.3631) | 13.7 | 0.6508 (0.0065) | 8.8 | 0.6519 (0.0063) | 9.0 | 6.0326 (6.3631) | 13.7 |
| | LEAP (PP-MLP) | 0.3070 (0.0132) | 5.6 | 0.3205 (0.0147) | 6.3 | 2.6398 (1.5599) | 13.2 | 0.6559 (0.0063) | 4.9 | 0.6570 (0.0061) | 5.9 | 2.6398 (1.5599) | 13.2 |
| | PP-LGBM | 0.3021 (0.0156) | 8.4 | 0.3175 (0.0165) | 8.3 | 0.0650 (0.0008) | 4.4 | 0.6562 (0.0055) | 4.8 | 0.6590 (0.0051) | 4.3 | 0.0650 (0.0008) | 4.4 |
| | LEAP (PP-LGBM) | 0.3089 (0.0156) | 4.8 | 0.3243 (0.0169) | 4.5 | 0.0647 (0.0008) | 2.5 | 0.6579 (0.0054) | 2.7 | 0.6606 (0.0050) | 2.5 | 0.0647 (0.0008) | 2.5 |
| 1 | PS-KNN | 0.2346 (0.0111) | 14.0 | 0.2505 (0.0118) | 14.0 | 0.0740 (0.0010) | 10.0 | 0.6083 (0.0055) | 12.0 | 0.6117 (0.0052) | 12.0 | 0.0740 (0.0010) | 10.0 |
| | LEAP (PS-KNN) | 0.2398 (0.0116) | 13.0 | 0.2561 (0.0122) | 13.0 | 0.0730 (0.0010) | 9.0 | 0.6129 (0.0058) | 11.0 | 0.6163 (0.0055) | 11.0 | 0.0730 (0.0010) | 9.0 |
| | PS-MLP | 0.3079 (0.0126) | 5.3 | 0.3223 (0.0131) | 5.2 | 0.8603 (0.0503) | 12.1 | 0.2572 (0.0133) | 13.9 | 0.1741 (0.0132) | 14.0 | 0.8603 (0.0503) | 12.1 |
| | LEAP (PS-MLP) | 0.3166 (0.0142) | 2.0 | 0.3316 (0.0149) | 2.1 | 0.5534 (0.0257) | 11.0 | 0.2629 (0.0150) | 13.1 | 0.2106 (0.0146) | 13.0 | 0.5534 (0.0257) | 11.0 |
| | PS-LGBM | 0.3028 (0.0141) | 8.2 | 0.3180 (0.0149) | 7.7 | 0.0650 (0.0007) | 4.5 | 0.6557 (0.0048) | 5.5 | 0.6585 (0.0043) | 4.9 | 0.0650 (0.0007) | 4.5 |
| | LEAP (PS-LGBM) | 0.3086 (0.0142) | 5.3 | 0.3240 (0.0150) | 4.4 | 0.0647 (0.0007) | 2.5 | 0.6579 (0.0048) | 2.6 | 0.6606 (0.0043) | 2.5 | 0.0647 (0.0007) | 2.5 |
| | PS-LASSO | 0.3094 (0.0138) | 4.6 | 0.3228 (0.0146) | 5.5 | 0.0657 (0.0009) | 6.0 | 0.6536 (0.0057) | 7.0 | 0.6552 (0.0052) | 7.0 | 0.0657 (0.0009) | 6.0 |
| | LEAP (PS-LASSO) | **0.3210 (0.0135)** | **1.0** | **0.3348 (0.0142)** | **1.0** | **0.0644 (0.0008)** | **1.1** | **0.6607 (0.0055)** | **1.0** | **0.6629 (0.0049)** | **1.1** | **0.0644 (0.0008)** | **1.1** |
| | PP-tDNN | 0.3295 (0.0146) | 11.7 | 0.3634 (0.0130) | 11.1 | 0.0088 (0.0003) | 6.8 | 0.7554 (0.0045) | 7.4 | 0.8107 (0.0041) | 7.9 | 0.0099 (0.0004) | 6.5 |
| | LEAP (PP-tDNN) | 0.3374 (0.0136) | 9.4 | 0.3708 (0.0118) | 8.9 | 0.0087 (0.0003) | 4.7 | 0.7579 (0.0044) | 4.5 | 0.8135 (0.0038) | 3.9 | 0.0098 (0.0004) | 4.4 |
| | PP-MLP | 0.3401 (0.0150) | 9.0 | 0.3704 (0.0114) | 10.1 | 7.9e5 (1.3e5) | 13.6 | 0.7562 (0.0048) | 6.8 | 0.8111 (0.0035) | 7.3 | 8.6e5 (1.4e6) | 13.6 |
| | LEAP (PP-MLP) | 0.3461 (0.0125) | 6.6 | 0.3763 (0.0111) | 6.9 | 3.2e5 (2.2e5) | 13.4 | 0.7583 (0.0045) | 3.9 | 0.8136 (0.0028) | 3.7 | 3.4e5 (2.4e5) | 13.4 |
| | PP-LGBM | 0.3600 (0.0092) | 2.4 | 0.3922 (0.0075) | 2.1 | 0.0086 (0.0003) | 2.8 | 0.7602 (0.0047) | 2.4 | 0.8142 (0.0035) | 3.3 | 0.0097 (0.0004) | 2.6 |
| | LEAP (PP-LGBM) | **0.3673 (0.0115)** | **1.0** | **0.3990 (0.0091)** | **1.0** | **0.0085 (0.0003)** | **1.1** | **0.7618 (0.0049)** | **1.1** | **0.8160 (0.0035)** | **1.6** | **0.0096 (0.0003)** | **1.2** |
| 2 | PS-KNN | 0.3019 (0.0113) | 14.0 | 0.3313 (0.0088) | 14.0 | 0.0097 (0.0003) | 10.0 | 0.7314 (0.0050) | 12.0 | 0.7885 (0.0040) | 12.0 | 0.0110 (0.0004) | 10.0 |
| | LEAP (PS-KNN) | 0.3061 (0.0117) | 13.0 | 0.3359 (0.0091) | 13.0 | 0.0096 (0.0003) | 9.0 | 0.7344 (0.0052) | 11.0 | 0.7910 (0.0040) | 11.0 | 0.0108 (0.0004) | 9.0 |
| | PS-MLP | 0.3499 (0.0109) | 5.2 | 0.3824 (0.0094) | 5.1 | 1.2242 (0.2137) | 12.0 | 0.1829 (0.0195) | 13.6 | 0.1681 (0.0144) | 14.0 | 1.1427 (0.1860) | 12.0 |
| | LEAP (PS-MLP) | 0.3570 (0.0112) | 2.8 | 0.3882 (0.0091) | 3.0 | 0.7205 (0.0944) | 11.0 | 0.1843 (0.0228) | 13.4 | 0.2140 (0.0133) | 13.0 | 0.6777 (0.0826) | 11.0 |
| | PS-LGBM | 0.3409 (0.0125) | 8.7 | 0.3730 (0.0086) | 8.7 | 0.0087 (0.0003) | 5.6 | 0.7568 (0.0048) | 6.0 | 0.8112 (0.0033) | 7.5 | 0.0098 (0.0003) | 6.0 |
| | LEAP (PS-LGBM) | 0.3472 (0.0132) | 6.4 | 0.3798 (0.0091) | 6.0 | 0.0087 (0.0003) | 3.7 | 0.7580 (0.0050) | 4.2 | 0.8125 (0.0034) | 5.3 | 0.0097 (0.0003) | 3.8 |
| | PS-LASSO | 0.3392 (0.0112) | 9.8 | 0.3707 (0.0094) | 9.7 | 0.0089 (0.0003) | 7.5 | 0.7467 (0.0054) | 10.0 | 0.8084 (0.0037) | 9.7 | 0.0099 (0.0003) | 7.5 |
| | LEAP (PS-LASSO) | 0.3502 (0.0113) | 5.0 | 0.3819 (0.0102) | 5.4 | 0.0086 (0.0003) | 3.8 | 0.7537 (0.0050) | 8.7 | 0.8129 (0.0035) | 4.8 | 0.0097 (0.0003) | 4.0 |
| | PP-tDNN | 0.1784 (0.0264) | 10.9 | 0.2021 (0.0348) | 9.6 | 0.0219 (0.0008) | 6.6 | 0.6333 (0.0099) | 6.9 | 0.6165 (0.0098) | 8.6 | 0.0215 (0.0007) | 7.0 |
| | LEAP (PP-tDNN) | 0.1837 (0.0246) | 8.5 | 0.2077 (0.0336) | 7.3 | 0.0218 (0.0007) | 6.1 | 0.6354 (0.0081) | 6.3 | 0.6188 (0.0080) | 7.8 | 0.0214 (0.0006) | 6.3 |
| | PP-MLP | 0.1756 (0.0286) | 11.1 | 0.1969 (0.0371) | 11.0 | 8.9e5 (1.5e6) | 14.6 | 0.6283 (0.0087) | 9.4 | 0.6154 (0.0091) | 9.3 | 8.7e5 (1.4e6) | 14.6 |
| | LEAP (PP-MLP) | 0.1860 (0.0307) | 7.8 | 0.2071 (0.0383) | 7.1 | 3.6e5 (2.5e5) | 14.4 | 0.6311 (0.0089) | 8.0 | 0.6186 (0.0090) | 7.5 | 3.5e5 (2.4e5) | 14.4 |
| | PP-LGBM | 0.1854 (0.0150) | 9.0 | 0.2027 (0.0201) | 9.6 | **0.0210 (0.0005)** | **1.3** | 0.6457 (0.0060) | 2.5 | 0.6335 (0.0061) | 2.5 | 0.0207 (0.0005) | 2.4 |
| | LEAP (PP-LGBM) | 0.1956 (0.0147) | 4.8 | 0.2133 (0.0197) | 5.0 | **0.0210 (0.0005)** | 1.7 | **0.6467 (0.0059)** | **1.2** | **0.6354 (0.0064)** | **1.0** | **0.0206 (0.0005)** | **1.0** |
| | PS-KNN | 0.1377 (0.0135) | 14.7 | 0.1516 (0.0156) | 14.9 | 0.0234 (0.0006) | 10.0 | 0.6080 (0.0079) | 12.0 | 0.5981 (0.0080) | 12.0 | 0.0229 (0.0005) | 10.0 |
| 2-PRISM | LEAP (PS-KNN) | 0.1428 (0.0121) | 13.4 | 0.1560 (0.0146) | 13.7 | 0.0232 (0.0006) | 9.0 | 0.6114 (0.0079) | 11.0 | 0.6009 (0.0081) | 10.9 | 0.0227 (0.0005) | 9.0 |
| | PS-MLP | 0.1937 (0.0199) | 6.1 | 0.2101 (0.0260) | 5.8 | 0.7251 (0.1247) | 13.0 | 0.0422 (0.0222) | 14.0 | 0.0235 (0.0223) | 14.5 | 0.7311 (0.1204) | 13.0 |
| | LEAP (PS-MLP) | 0.1989 (0.0187) | 3.7 | 0.2157 (0.0245) | 3.6 | 0.4423 (0.0363) | 12.0 | 0.0163 (0.0256) | 15.0 | 0.0167 (0.0227) | 14.5 | 0.4417 (0.0328) | 12.0 |
| | PS-LGBM | 0.1822 (0.0116) | 10.2 | 0.1998 (0.0194) | 10.7 | 0.0213 (0.0005) | 3.3 | 0.6446 (0.0053) | 3.8 | 0.6317 (0.0067) | 4.0 | 0.0208 (0.0005) | 4.0 |
| | LEAP (PS-LGBM) | 0.1903 (0.0122) | 7.5 | 0.2084 (0.0200) | 7.4 | 0.0213 (0.0005) | 3.8 | 0.6455 (0.0051) | 2.6 | 0.6331 (0.0062) | 2.5 | 0.0207 (0.0004) | 2.6 |
| | PS-LASSO | 0.1910 (0.0207) | 7.1 | 0.2050 (0.0242) | 8.2 | 0.0221 (0.0005) | 7.3 | 0.6311 (0.0066) | 8.4 | 0.6223 (0.0078) | 6.6 | 0.0215 (0.0004) | 7.0 |
| | LEAP (PS-LASSO) | 0.1989 (0.0217) | 3.7 | 0.2137 (0.0253) | 4.5 | 0.0218 (0.0005) | 5.9 | 0.6359 (0.0063) | 5.9 | 0.6266 (0.0071) | 5.3 | 0.0212 (0.0004) | 5.7 |
| | Reproducibility | **0.2200 (0.0188)** | **1.5** | **0.2454 (0.0258)** | **1.6** | 0.0277 (0.0008) | 11.0 | 0.5547 (0.0098) | 13.0 | 0.5647 (0.0075) | 13.0 | 0.0272 (0.0008) | 11.0 |
| | PS-KNN | 0.1109 (0.0006) | 8.0 | 0.1212 (0.0006) | 8.0 | 0.0782 (0.0001) | 6.0 | 0.5841 (0.0004) | 6.0 | 0.5883 (0.0004) | 6.0 | 0.0782 (0.0001) | 6.0 |
| | LEAP (PS-KNN) | 0.1158 (0.0006) | 7.0 | 0.1266 (0.0006) | 7.0 | 0.0770 (0.0001) | 5.0 | 0.5897 (0.0004) | 5.0 | 0.5939 (0.0004) | 5.0 | 0.0770 (0.0001) | 5.0 |
| | PS-MLP | 0.1935 (0.0006) | 4.0 | 0.2055 (0.0006) | 4.0 | 0.8353 (0.0012) | 8.0 | 0.2195 (0.0005) | 8.0 | 0.1459 (0.0005) | 8.0 | 0.8353 (0.0012) | 8.0 |
| 3 | LEAP (PS-MLP) | 0.2029 (0.0006) | 2.0 | 0.2158 (0.0006) | 2.0 | 0.5560 (0.0006) | 7.0 | 0.2233 (0.0004) | 7.0 | 0.1764 (0.0005) | 7.0 | 0.5560 (0.0006) | 7.0 |
| | PS-LGBM | 0.1754 (0.0005) | 6.0 | 0.1902 (0.0005) | 6.0 | 0.0676 (0.0000) | 2.0 | 0.6385 (0.0003) | 2.0 | 0.6421 (0.0003) | 2.0 | 0.0676 (0.0000) | 2.0 |
| | LEAP (PS-LGBM) | 0.1843 (0.0006) | 5.0 | 0.1989 (0.0006) | 5.0 | **0.0674 (0.0000)** | **1.0** | **0.6407 (0.0003)** | **1.0** | **0.6441 (0.0003)** | **1.0** | **0.0674 (0.0000)** | **1.0** |
| | PS-LASSO | 0.1981 (0.0005) | 3.0 | 0.2088 (0.0005) | 3.0 | 0.0712 (0.0001) | 4.0 | 0.6271 (0.0003) | 4.0 | 0.6258 (0.0003) | 4.0 | 0.0712 (0.0001) | 4.0 |
| | LEAP (PS-LASSO) | **0.2090 (0.0005)** | **1.0** | **0.2200 (0.0005)** | **1.0** | 0.0690 (0.0001) | 3.0 | 0.6359 (0.0003) | 3.0 | 0.6370 (0.0003) | 3.0 | 0.0690 (0.0001) | 3.0 |
| | PS-KNN | 0.0813 (0.0011) | 8.0 | 0.0919 (0.0011) | 8.0 | 0.0126 (0.0000) | 4.9 | 0.6877 (0.0004) | 6.0 | 0.7298 (0.0005) | 4.9 | 0.0139 (0.0000) | 4.9 |
| | LEAP (PS-KNN) | 0.0826 (0.0011) | 7.0 | 0.0942 (0.0011) | 7.0 | 0.0124 (0.0000) | 3.9 | 0.6916 (0.0004) | 4.6 | 0.7332 (0.0005) | 3.9 | 0.0137 (0.0000) | 3.9 |
| | PS-MLP | 0.1536 (0.0010) | 2.2 | 0.1675 (0.0011) | 2.0 | 0.9169 (0.0027) | 8.0 | 0.1201 (0.0008) | 8.0 | 0.1013 (0.0007) | 8.0 | 0.8329 (0.0023) | 8.0 |
| 4 | LEAP (PS-MLP) | **0.1599 (0.0009)** | **1.0** | **0.1744 (0.0010)** | **1.0** | 0.5852 (0.0012) | 7.0 | 0.1220 (0.0007) | 7.0 | 0.1208 (0.0008) | 7.0 | 0.5354 (0.0010) | 7.0 |
| | PS-LGBM | 0.1314 (0.0010) | 6.0 | 0.1493 (0.0009) | 6.0 | 0.0108 (0.0000) | 2.9 | 0.7277 (0.0004) | 2.0 | 0.7682 (0.0004) | 2.0 | 0.0119 (0.0000) | 2.9 |
| | LEAP (PS-LGBM) | 0.1392 (0.0009) | 5.0 | 0.1567 (0.0009) | 4.7 | 0.0108 (0.0000) | 1.9 | **0.7289 (0.0004)** | **1.0** | **0.7691 (0.0004)** | **1.0** | 0.0119 (0.0000) | 1.9 |
| | PS-LASSO | 0.1469 (0.0033) | 4.0 | 0.1577 (0.0030) | 4.3 | **0.0105 (0.0013)** | **1.5** | 0.6917 (0.0017) | 4.4 | 0.7430 (0.0085) | 3.3 | **0.0116 (0.0016)** | **1.5** |
| | LEAP (PS-LASSO) | 0.1528 (0.0009) | 2.9 | 0.1648 (0.0010) | 3.0 | 0.0134 (0.0000) | 5.9 | 0.6994 (0.0004) | 3.0 | 0.7291 (0.0005) | 5.9 | 0.0153 (0.0000) | 5.9 |

**Supplementary Table 3. Performances in predicting the response to 5 drugs in PDX samples (task 5).** We train our novel approach (LEAP) to predict the AUC in cell lines from the GDSC study. We evaluate performances in PDX using the Spearman's and Pearson's correlations between predicted and observed response to each of the drugs. To assess variability we report the mean (standard deviation) of these metrics calculated over 100 subsets of the data obtained by removing 10 PDX samples.

| Drug | Model | Spearman's correlation | Pearson's correlation |
|---|---|---|---|
| Cetuximab | PS-LASSO | **0.3203 (0.0475)** | **0.3397 (0.0508)** |
| | LEAP (PS-LASSO) | 0.3018 (0.0482) | 0.3324 (0.0506) |
| Erlotinib | PS-LASSO | 0.5349 (0.1259) | 0.5499 (0.1436) |
| | LEAP (PS-LASSO) | **0.5619 (0.1215)** | **0.5943 (0.1232)** |
| Tamoxifen | PS-LASSO | -0.2639 (0.1041) | -0.1603 (0.1011) |
| | LEAP (PS-LASSO) | **-0.0275 (0.1060)** | **-0.0804 (0.1068)** |
| Trametinib | PS-LASSO | 0.2257 (0.1104) | 0.4540 (0.1107) |
| | LEAP (PS-LASSO) | **0.2966 (0.1036)** | **0.4596 (0.1118)** |
| Paclitaxel | PS-LASSO | **-0.0555 (0.0629)** | **-0.1548 (0.0565)** |
| | LEAP (PS-LASSO) | -0.1032 (0.0643) | -0.1953 (0.0550) |

**Supplementary Table 4. Ablation study quantifying the change in prediction performance due to each of the methodological choices on tasks 1 and 2.** Performances are reported on 10 different held out test sets and represent the average performance per perturbation. In step 0, we use the PS-LASSO model with a unique preprocessor and MAE representation model trained on each of the 10 training sets. In step 1, we pre-train a single preprocessor and MAE model on all DepMap transcriptomics data that are later used to obtain the cell line representations in all trainings. In Step 2, the preprocessor and MAE are pretrained on a concatenation of transcriptomics data from DepMap and GDSC. In Step 3, transcriptomics from PDXE are added to the pretraining data. In Step 4, 5 PS-LASSO prediction heads trained in 5-fold cross-validation splits of the transformed gene expression representation are ensembled. Finally, in step 5, we repeat the previous ensemble training of the 5 PS-LASSO models using 5 different MAE (where initialisation varies) representations, leading to the LEAP (PS-LASSO) model.

| Task | Ablation study step | Dataset for preprocessing and representation model training | Number of representation models for ensembling | Total number of prediction models for ensembling | Spearman's correlation |
|---|---|---|---|---|---|
| 1 | 0 | Training set | 1 | 1 | 0.3095 (0.0141) |
| | 1 | **Full DepMap (version 23Q4)** | 1 | 1 | 0.3057 (0.0135) |
| | 2 | **Full DepMap (version 23Q4) and GDSC** | 1 | 1 | 0.3059 (0.0133) |
| | 3 | **Full DepMap (version 23Q4), GDSC and PDX Encyclopedia** | 1 | 1 | 0.3063 (0.0136) |
| | 4 | Full DepMap (version 23Q4), GDSC and PDX Encyclopedia | 1 | **5** | 0.3094 (0.0138) |
| | 5 | Full DepMap (version 23Q4), GDSC and PDX Encyclopedia | **5** | **25** | 0.3210 (0.0135) |
| 2 | 0 | Training set | 1 | 1 | 0.3368 (0.0114) |
| | 1 | **Full DepMap (version 23Q4)** | 1 | 1 | 0.3333 (0.0123) |
| | 2 | **Full DepMap (version 23Q4) and GDSC** | 1 | 1 | 0.3344 (0.0103) |
| | 3 | **Full DepMap (version 23Q4), GDSC and PDX Encyclopedia** | 1 | 1 | 0.3354 (0.0109) |
| | 4 | Full DepMap (version 23Q4), GDSC and PDX Encyclopedia | 1 | **5** | 0.3392 (0.0112) |
| | 5 | Full DepMap (version 23Q4), GDSC and PDX Encyclopedia | **5** | **25** | 0.3502 (0.0113) |

**Supplementary Table 5. Performances in predicting gene essentiality (task 1) and drug response (task 2) in unseen cell lines using different ensembling strategies.** We compare the prediction performances with PS-LASSO prediction models in three different ways: using ensembling over the 5-fold cross-validation (5 models), using ensembling over 5 repeats of the 5-fold cross-validation (25 models), or using LEAP (25 models). Performances are evaluated over 10 splits of the data into non-overlapping training and test sets. For each training-test split, we calculate the average Spearman's correlations, Pearson's correlations and Mean Squared Error comparing observed and predicted responses per perturbation. We report the mean (standard deviation) of these metrics calculated over the training-test splits.

| Task | Approach | Spearman's correlation | Pearson's correlation | Mean Squared Error |
|---|---|---|---|---|
| | PS-LASSO (ensemble of 5 models) | 0.3094 (0.0138) | 0.3228 (0.0146) | 0.0657 (0.0009) |
| 1 | PS-LASSO (ensemble of 25 models) | 0.3141 (0.0138) | 0.3276 (0.0147) | 0.0651 (0.0009) |
| | PS-LASSO (LEAP) | **0.3210 (0.0135)** | **0.3348 (0.0142)** | **0.0644 (0.0008)** |
| | PS-LASSO (ensemble of 5 models) | 0.3392 (0.0112) | 0.3707 (0.0094) | 0.0089 (0.0003) |
| 2 | PS-LASSO (ensemble of 25 models) | 0.3436 (0.0110) | 0.3753 (0.0095) | 0.0088 (0.0003) |
| | PS-LASSO (LEAP) | **0.3502 (0.0113)** | **0.3819 (0.0102)** | **0.0086 (0.0003)** |

**Supplementary Table 6. Performances in predicting gene dependency in unseen cell lines (task 1').** We compare the performance of our novel approach (LEAP) with previously reported performances obtained using different representation models in OmicsRPZ, and DeepDEP. Performances are evaluated over 10 splits of the data into non-overlapping sets of cell lines. For each data split, we report the average Spearman's correlation between observed and predicted gene dependencies calculated over all 1,223 dependencies of interest. For each training-test split, we calculate the mean perturbation-specific Spearman's correlations comparing observed and predicted gene dependencies. We report the mean (standard deviation) of this metrics calculated over the 10 training-test splits. Note that the performances of OmicsRPZ models and DeepDEP were not recomputed and were obtained with different training-test splits of the same dataset.

| Approach | Spearman's correlation |
|---|---|
| DeepDEP | 0.1732 (0.0160) |
| OmicsRPZ (Identity) | 0.2369 (0.0058) |
| OmicsRPZ (scVI) | 0.2452 (0.0139) |
| OmicsRPZ (PCA) | 0.2461 (0.0055) |
| OmicsRPZ (MHAE) | 0.2489 (0.0032) |
| OmicsRPZ (AE) | 0.2489 (0.0078) |
| OmicsRPZ (GNN) | 0.2502 (0.0095) |
| OmicsRPZ (MAE) | 0.2551 (0.0059) |
| OmicsRPZ (DA-GN) | 0.2559 (0.0059) |
| PS-LASSO | 0.2856 (0.0054) |
| LEAP (PS_LASSO) | 0.2972 (0.0056) |

**Supplementary Table 7. Stability selection to identify important Reactome pathways for gene essentiality prediction in LEAP (PS-LASSO).** For top 20 best predicted genes in task 1, we report the prediction performance (Spearman's correlation) obtained with LEAP (PS-LASSO), the normalised stability score obtained with the tuned models, and the number and list of stably selected Reactome pathways (annotated with their identifiers or pathway names).

| Gene | Prediction performance with LEAP (PS-LASSO) | Stability score | Number of selected pathways | Selected pathway identifiers | Selected pathway names |
|---|---|---|---|---|---|
| FERMT2 | 0.785 | 0.89 | 6 | R-HSA-190370, R-HSA-2024101, R-HSA-210990, R-HSA-3000178, R-HSA-3299685, R-HSA-9619665 | FGFR1b ligand binding and activation, CS/DS degradation, PECAM1 interactions, ECM proteoglycans, Detoxification of Reactive Oxygen Species, EGR2 and SOX10-mediated initiation of Schwann cell myelination |
| ITGAV | 0.739 | 0.85 | 9 | R-HSA-114608, R-HSA-190370, R-HSA-190373, R-HSA-210990, R-HSA-3000178, R-HSA-3299685, R-HSA-5658623, R-HSA-9619665, R-HSA-9840310 | Platelet degranulation, FGFR1b ligand binding and activation, FGFR1c ligand binding and activation, PECAM1 interactions, ECM proteoglycans, Detoxification of Reactive Oxygen Species, FGFRL1 modulation of FGFR1 signaling, EGR2 and SOX10-mediated initiation of Schwann cell myelination, Glycosphingolipid catabolism |
| CDS2 | 0.718 | 0.95 | 4 | R-HSA-2173791, R-HSA-351906, R-HSA-420029, R-HSA-6806942 | TGF-beta receptor signaling in EMT (epithelial to mesenchymal transition), Apoptotic cleavage of cell adhesion proteins, Tight junction interactions, MET Receptor Activation |
| EFR3A | 0.677 | 0.86 | 10 | R-HSA-163615, R-HSA-164378, R-HSA-170660, R-HSA-389977, R-HSA-448706, R-HSA-5620971, R-HSA-5635838, R-HSA-75876, R-HSA-83936, R-HSA-9609736 | PKA activation, PKA activation in glucagon signalling, Adenylate cyclase activating pathway, Post-chaperonin tubulin folding pathway, Interleukin-1 processing, Pyroptosis, Activation of SMO, Synthesis of very long-chain fatty acyl-CoAs, Transport of nucleosides and free purine and pyrimidine bases across the plasma membrane, Assembly and cell surface presentation of NMDA receptors |
| STX4 | 0.656 | 0.83 | 19 | R-HSA-1483115, R-HSA-156584, R-HSA-1606341, R-HSA-1839120, R-HSA-190370, R-HSA-190377, R-HSA-211935, R-HSA-2142770, R-HSA-3296197, R-HSA-389977, R-HSA-418597, R-HSA-425381, R-HSA-5250971, R-HSA-5250992, R-HSA-6803205, R-HSA-8939247, R-HSA-9609736, R-HSA-9733458, R-HSA-9758919 | Hydrolysis of LPC, Cytosolic sulfonation of small molecules, IRF3 mediated activation of type 1 IFN, Signaling by FGFR1 amplification mutants, FGFR1b ligand binding and activation, FGFR2b ligand binding and activation, Fatty acids, Synthesis of 15-eicosatetraenoic acid derivatives, Hydroxycarboxylic acid-binding receptors, Post-chaperonin tubulin folding pathway, G alpha (z) signalling events, Bicarbonate transporters, Toxicity of botulinum toxin type C (botC), Toxicity of botulinum toxin type E (botE), TP53 regulates transcription of several additional cell death genes whose specific roles in p53-dependent apoptosis remain uncertain, RUNX1 regulates transcription of genes involved in interleukin signaling, Assembly and cell surface presentation of NMDA receptors, Induction of Cell-Cell Fusion, Epithelial-Mesenchymal Transition (EMT) during gastrulation |
| CHMP4B | 0.648 | 0.91 | 5 | R-HSA-5621480, R-HSA-8852405, R-HSA-9037629, R-HSA-9758919, R-HSA-9840309 | Dectin-2 family, Signaling by MST1, Lewis blood group biosynthesis, Epithelial-Mesenchymal Transition (EMT) during gastrulation, Glycosphingolipid biosynthesis |
| MYBL2 | 0.647 | 0.78 | 11 | R-HSA-1483115, R-HSA-1839122, R-HSA-190372, R-HSA-190377, R-HSA-3295583, R-HSA-5221030, R-HSA-5576890, R-HSA-6803544, R-HSA-9840310, R-HSA-9854907, R-HSA-9857377 | Hydrolysis of LPC, Signaling by activated point mutants of FGFR1, FGFR3c ligand binding and activation, FGFR2b ligand binding and activation, TRP channels, TET1,2,3 and TDG demethylate DNA, Phase 3 - rapid repolarisation, Ion influx/efflux at host-pathogen interface, Glycosphingolipid catabolism, Regulation of MITF-M dependent genes involved in metabolism, Regulation of MITF-M-dependent genes involved in lysosome biogenesis and autophagy |
| CCND1 | 0.633 | 0.79 | 27 | R-HSA-111453, R-HSA-1236973, R-HSA-159418, R-HSA-2485179, R-HSA-352238, R-HSA-379398, R-HSA-390522, R-HSA-416700, R-HSA-449836, R-HSA-4549380, R-HSA-5619066, R-HSA-5619078, R-HSA-8951430, R-HSA-8952158, R-HSA-936837, R-HSA-9630794, R-HSA-9632700, R-HSA-9634638, R-HSA-9636383, R-HSA-9646303, R-HSA-9646304, R-HSA-9661069, R-HSA-9680350, R-HSA-9706374, R-HSA-9754119, R-HSA-9823730, R-HSA-9827857 | BH3-only proteins associate with and inactivate anti-apoptotic BCL-2 members, Cross-presentation of particulate exogenous antigens (phagosomes), Recycling of bile acids and salts, Activation of the phototransduction cascade, Breakdown of the nuclear lamina, Enzymatic degradation of Dopamine by monoamine oxidase, Striated Muscle Contraction, Other semaphorin interactions, Other interleukin signaling, Defective ALG1 causes CDG-1k, Defective SLC22A18 causes lung cancer (LNCR) and embryonal rhabdomyosarcoma 1 (RMSE1), Defective SLC35C1 causes congenital disorder of glycosylation 2C (CDG2C), RUNX3 regulates WNT signaling, RUNX3 regulates BCL2L11 (BIM) transcription, Ion transport by P-type ATPases, Evasion of Oncogene Induced Senescence Due to Defective p16INK4A binding to CDK4 and CDK6, Evasion of Oxidative Stress Induced Senescence Due to Defective p16INK4A binding to CDK4 and CDK6, Estrogen-dependent nuclear events downstream of ESR-membrane signaling, Prevention of phagosomal-lysosomal fusion, Evasion of Oncogene Induced Senescence Due to p14ARF Defects, Evasion of Oxidative Stress Induced Senescence Due to p14ARF Defects, Defective binding of RB1 mutants to E2F1,(E2F2, E2F3), Signaling by CSF1 (M-CSF) in myeloid cells, FLT3 signaling through SRC family kinases, Drug-mediated inhibition of CDK4/CDK6 activity, Formation of definitive endoderm, Specification of primordial germ cells |
| PTK2 | 0.631 | 0.85 | 17 | R-HSA-1592389, R-HSA-190373, R-HSA-210990, R-HSA-216083, R-HSA-3299685, R-HSA-3560783, R-HSA-381426, R-HSA-391160, R-HSA-416700, R-HSA-428359, R-HSA-446343, R-HSA-5578996, R-HSA-5658623, R-HSA-8939246, R-HSA-8964315, R-HSA-9619665, R-HSA-9761174 | Activation of Matrix Metalloproteinases, FGFR1c ligand binding and activation, PECAM1 interactions, Integrin cell surface interactions, Detoxification of Reactive Oxygen Species, Defective B4GALT7 causes EDS, progeroid type, Regulation of Insulin-like Growth Factor (IGF) transport and uptake by Insulin-like Growth Factor Binding Proteins (IGFBPs), Signal regulatory protein family interactions, Other semaphorin interactions, Insulin-like Growth Factor-2 mRNA Binding Proteins (IGF2BPs/IMPs/VICKZs) bind RNA, Localization of the PINCH-ILK-PARVIN complex to focal adhesions, Defective CYP27A1 causes CTX, FGFRL1 modulation of FGFR1 signaling, RUNX1 regulates transcription of genes involved in differentiation of myeloid cells, G beta:gamma signalling through BTK, EGR2 and SOX10-mediated initiation of Schwann cell myelination, Formation of intermediate mesoderm |
| SDHB | 0.624 | 0.89 | 12 | R-HSA-190704, R-HSA-190827, R-HSA-191650, R-HSA-2129379, R-HSA-216083, R-HSA-2214320, R-HSA-2243919, R-HSA-2871809, R-HSA-419408, R-HSA-6785807, R-HSA-8874081, R-HSA-983695 | Oligomerization of connexins into connexons, Transport of connexins along the secretory pathway, Regulation of gap junction activity, Molecules associated with elastic fibres, Integrin cell surface interactions, Anchoring fibril formation, Crosslinking of collagen fibrils, FCERI mediated Ca+2 mobilization, Lysosphingolipid and LPA receptors, Interleukin-4 and Interleukin-13 signaling, MET activates PTK2 signaling, Antigen activates B Cell Receptor (BCR) leading to generation of second messengers |
| TLN1 | 0.623 | 0.92 | 6 | R-HSA-114608, R-HSA-3000157, R-HSA-446343, R-HSA-6785807, R-HSA-6788467, R-HSA-75205 | Platelet degranulation, Laminin interactions, Localization of the PINCH-ILK-PARVIN complex to focal adhesions, Interleukin-4 and Interleukin-13 signaling, IL-6-type cytokine receptor ligand interactions, Dissolution of Fibrin Clot |
| ELMO2 | 0.619 | 0.91 | 15 | R-HSA-114516, R-HSA-1483115, R-HSA-1566977, R-HSA-1839122, R-HSA-190372, R-HSA-190373, R-HSA-193634, R-HSA-196025, R-HSA-3560796, R-HSA-445355, R-HSA-446343, R-HSA-5654219, R-HSA-5676934, R-HSA-9033807, R-HSA-9761174 | Disinhibition of SNARE formation, Hydrolysis of LPC, Fibronectin matrix formation, Signaling by activated point mutants of FGFR1, FGFR3c ligand binding and activation, FGFR1c ligand binding and activation, Axonal growth inhibition (RHOA activation), Formation of annular gap junctions, Defective PAPSS2 causes SEMD-PA, Smooth Muscle Contraction, Localization of the PINCH-ILK-PARVIN complex to focal adhesions, Phospholipase C-mediated cascade: FGFR1, Protein repair, ABO blood group biosynthesis, Formation of intermediate mesoderm |
| RAB6A | 0.615 | 0.82 | 11 | R-HSA-164378, R-HSA-170660, R-HSA-1839120, R-HSA-210990, R-HSA-3295583, R-HSA-388844, R-HSA-389977, R-HSA-427601, R-HSA-5635838, R-HSA-75876, R-HSA-975578 | PKA activation in glucagon signalling, Adenylate cyclase activating pathway, Signaling by FGFR1 amplification mutants, PECAM1 interactions, TRP channels, Receptor-type tyrosine-protein phosphatases, Post-chaperonin tubulin folding pathway, Multifunctional anion exchangers, Activation of SMO, Synthesis of very long-chain fatty acyl-CoAs, Reactions specific to the complex N-glycan synthesis pathway |
| WWTR1 | 0.615 | 0.89 | 8 | R-HSA-114608, R-HSA-216083, R-HSA-3000157, R-HSA-381426, R-HSA-5654719, R-HSA-6785807, R-HSA-6788467, R-HSA-75205 | Platelet degranulation, Integrin cell surface interactions, Laminin interactions, Regulation of Insulin-like Growth Factor (IGF) transport and uptake by Insulin-like Growth Factor Binding Proteins (IGFBPs), SHC-mediated cascade:FGFR4, Interleukin-4 and Interleukin-13 signaling, IL-6-type cytokine receptor ligand interactions, Dissolution of Fibrin Clot |
| ITGB5 | 0.615 | 0.75 | 73 | R-HSA-114516, R-HSA-114604, R-HSA-140180, R-HSA-1483115, R-HSA-1839120, R-HSA-190377, R-HSA-190704, R-HSA-190827, R-HSA-193634, R-HSA-196025, R-HSA-2022377, R-HSA-2023837, R-HSA-209822, R-HSA-2142670, R-HSA-2168880, R-HSA-2206307, R-HSA-3296197, R-HSA-350864, R-HSA-3560796, R-HSA-376172, R-HSA-389887, R-HSA-389948, R-HSA-416550, R-HSA-425381, R-HSA-4341670, R-HSA-446343, R-HSA-4724289, R-HSA-5083629, R-HSA-5083633, R-HSA-5579011, R-HSA-5579012, R-HSA-5579013, R-HSA-5579014, R-HSA-5619039, R-HSA-5619047, R-HSA-5619052, R-HSA-5619053, R-HSA-5619061, R-HSA-5626978, R-HSA-5682113, R-HSA-5684045, R-HSA-5690338, R-HSA-6806664, R-HSA-83936, R-HSA-844455, R-HSA-844615, R-HSA-8847453, R-HSA-8853333, R-HSA-8951671, R-HSA-9014826, R-HSA-9014843, R-HSA-9020265, R-HSA-9020933, R-HSA-9027604, R-HSA-9031528, R-HSA-9032759, R-HSA-9033807, R-HSA-9608287, R-HSA-9608290, R-HSA-9613354, R-HSA-9648895, R-HSA-9652817, R-HSA-9656255, R-HSA-9656256, R-HSA-9657050, R-HSA-9706374, R-HSA-9717264, R-HSA-9717301, R-HSA-9717316, R-HSA-9717319, R-HSA-9735763, R-HSA-975574, R-HSA-9758881 | Disinhibition of SNARE formation, GPVI-mediated activation cascade, COX reactions, Hydrolysis of LPC, Signaling by FGFR1 amplification mutants, FGFR2b ligand binding and activation, Oligomerization of connexins into connexons, Transport of connexins along the secretory pathway, Axonal growth inhibition (RHOA activation), Formation of annular gap junctions, Metabolism of Angiotensinogen to Angiotensins, Signaling by FGFR2 amplification mutants, Glycoprotein hormones, Synthesis of epoxy (EET) and dihydroxyeicosatrienoic acids (DHET), Scavenging of heme from plasma, MPS IIIA - Sanfilippo syndrome A, Hydroxycarboxylic acid-binding receptors, Regulation of thyroid hormone activity, Defective PAPSS2 causes SEMD-PA, DSCAM interactions, Beta-oxidation of pristanoyl-CoA, PD-1 signaling, Sema4D mediated inhibition of cell attachment and migration, Bicarbonate transporters, Defective NEU1 causes sialidosis, Localization of the PINCH-ILK-PARVIN complex to focal adhesions, Defective ALG6 causes CDG-1c, Defective POMT2 causes MDDGA2, MDDGB2 and MDDGC2, Defective POMT1 causes MDDGA1, MDDGB1 and MDDGC1, Defective CYP2U1 causes SPG56, Defective MAOA causes BRUNS, Defective CYP7B1 causes SPG5A and CBAS3, Defective CYP27B1 causes VDDR1A, Defective SLC12A6 causes agenesis of the corpus callosum, with peripheral neuropathy (ACCPN), Defective SLC2A9 causes hypouricemia renal 2 (RHUC2), Defective SLC9A9 causes autism 16 (AUTS16), Defective SLC22A5 causes systemic primary carnitine deficiency (CDSP), Defective SLC33A1 causes spastic paraplegia 42 (SPG42), TNFR1-mediated ceramide production, Defective ABCA1 causes TGD, Defective ABCD1 causes ALD, Defective ABCC6 causes PXE, Metabolism of vitamin K, Transport of nucleosides and free purine and pyrimidine bases across the plasma membrane, The NLRP1 inflammasome, The AIM2 inflammasome, Synthesis of PIPs in the nucleus, Signaling by FGFR2 fusions, RUNX3 regulates YAP1-mediated transcription, Interleukin-36 pathway, Interleukin-33 signaling, Biosynthesis of aspirin-triggered D-series resolvins, Interleukin-23 signaling, Biosynthesis of electrophilic ω-3 PUFA oxo-derivatives, NR1H2 & NR1H3 regulate gene expression linked to triglyceride lipolysis in adipose, NTRK2 activates RAC1, ABO blood group biosynthesis, Defective MUTYH substrate binding, Defective MUTYH substrate processing, Lipophagy, Response of EIF2AK1 (HRI) to heme deficiency, Signaling by MAPK mutants, Defective OGG1 Substrate Binding, Defective OGG1 Substrate Processing, Defective OGG1 Localization, FLT3 signaling through SRC family kinases, ASP-3026-resistant ALK mutants, NVP-TAE684-resistant ALK mutants, alectinib-resistant ALK mutants, brigatinib-resistant ALK mutants, Defective PNP disrupts phosphorolysis of (deoxy)guanosine and (deoxy)inosine, Reactions specific to the hybrid N-glycan synthesis pathway, Uptake of dietary cobalamins into enterocytes |
| MDM2 | 0.606 | 0.83 | 14 | R-HSA-139915, R-HSA-174403, R-HSA-3296197, R-HSA-5619039, R-HSA-6803211, R-HSA-68952, R-HSA-69895, R-HSA-8981607, R-HSA-9022702, R-HSA-9603505, R-HSA-9645722, R-HSA-9646303, R-HSA-9646304, R-HSA-9764302 | Activation of PUMA and translocation to mitochondria, Glutathione synthesis and recycling, Hydroxycarboxylic acid-binding receptors, Defective SLC12A6 causes agenesis of the corpus callosum, with peripheral neuropathy (ACCPN), TP53 Regulates Transcription of Death Receptors and Ligands, DNA replication initiation, Transcriptional activation of cell cycle inhibitor p21, Intracellular oxygen transport, MECP2 regulates transcription of neuronal ligands, NTRK3 as a dependence receptor, Defective Intrinsic Pathway for Apoptosis Due to p14ARF Loss of Function, Evasion of Oncogene Induced Senescence Due to p14ARF Defects, Evasion of Oxidative Stress Induced Senescence Due to p14ARF Defects, Regulation of CDH19 Expression and Function |
| GPX4 | 0.603 | 0.88 | 33 | R-HSA-1306955, R-HSA-163358, R-HSA-164944, R-HSA-174403, R-HSA-189483, R-HSA-190704, R-HSA-190827, R-HSA-196108, R-HSA-2022923, R-HSA-2142712, R-HSA-351906, R-HSA-391908, R-HSA-5339717, R-HSA-5661270, R-HSA-5662702, R-HSA-5676594, R-HSA-5679001, R-HSA-6803204, R-HSA-6803205, R-HSA-71288, R-HSA-8849473, R-HSA-8866376, R-HSA-8866904, R-HSA-8964616, R-HSA-9029558, R-HSA-936837, R-HSA-9603381, R-HSA-9657689, R-HSA-9667769, R-HSA-9840309, R-HSA-9845619, R-HSA-9845621, R-HSA-9845622 | GRB7 events in ERBB2 signaling, PKA-mediated phosphorylation of key metabolic factors, Nef and signal transduction, Glutathione synthesis and recycling, Heme degradation, Oligomerization of connexins into connexons, Transport of connexins along the secretory pathway, Pregnenolone biosynthesis, Dermatan sulfate biosynthesis, Synthesis of 12-eicosatetraenoic acid derivatives, Apoptotic cleavage of cell adhesion proteins, Prostanoid ligand receptors, Signaling by LRP5 mutants, Formation of xylulose-5-phosphate, Melanin biosynthesis, TNF receptor superfamily (TNFSF) members mediating non-canonical NF-kB pathway, Defective ABCC2 causes DJS, TP53 Regulates Transcription of Genes Involved in Cytochrome C Release, TP53 regulates transcription of several additional cell death genes whose specific roles in p53-dependent apoptosis remain uncertain, Creatine metabolism, PTK6 Expression, Reelin signalling pathway, Negative regulation of activity of TFAP2 (AP-2) family transcription factors, G beta:gamma signalling through CDC42, NR1H2 & NR1H3 regulate gene expression linked to lipogenesis, Ion transport by P-type ATPases, Activated NTRK3 signals through PI3K, Defective SERPING1 causes hereditary angioedema, Acetylcholine inhibits contraction of outer hair cells, Glycosphingolipid biosynthesis, Enhanced cleavage of VWF variant by ADAMTS13, Defective VWF cleavage by ADAMTS13 variant, Defective VWF binding to collagen type I |
| ARHGEF7 | 0.602 | 0.83 | 4 | R-HSA-173736, R-HSA-180292, R-HSA-446107, R-HSA-9634638 | Alternative complement activation, GAB1 signalosome, Type I hemidesmosome assembly, Estrogen-dependent nuclear events downstream of ESR-membrane signaling |
| ZFP36L1 | 0.597 | 0.88 | 23 | R-HSA-1250347, R-HSA-1296067, R-HSA-165160, R-HSA-173736, R-HSA-190704, R-HSA-190827, R-HSA-196819, R-HSA-2214320, R-HSA-3238698, R-HSA-390650, R-HSA-444473, R-HSA-5619043, R-HSA-5655799, R-HSA-5657562, R-HSA-5661270, R-HSA-6788467, R-HSA-73614, R-HSA-75094, R-HSA-9617828, R-HSA-964827, R-HSA-9660826, R-HSA-9662834, R-HSA-9761174 | SHC1 events in ERBB4 signaling, Potassium transport channels, PDE3B signalling, Alternative complement activation, Oligomerization of connexins into connexons, Transport of connexins along the secretory pathway, Vitamin B1 (thiamin) metabolism, Anchoring fibril formation, WNT ligand biogenesis and trafficking, Histamine receptors, Formyl peptide receptors bind formyl peptides and many other ligands, Defective SLC2A1 causes GLUT1 deficiency syndrome 1 (GLUT1DS1), Defective SLC40A1 causes hemochromatosis 4 (HFE4) (duodenum), Essential fructosuria, Formation of xylulose-5-phosphate, IL-6-type cytokine receptor ligand interactions, Pyrimidine salvage, Formation of the Editosome, FOXO-mediated transcription of cell cycle genes, Progressive trimming of alpha-1,2-linked mannose residues from Man9/8/7GlcNAc2 to produce Man5GlcNAc2, Purinergic signaling in leishmaniasis infection, CD163 mediating an anti-inflammatory response, Formation of intermediate mesoderm |
| PFAS | 0.597 | 0.90 | 4 | R-HSA-216083, R-HSA-3000178, R-HSA-419408, R-HSA-6785807 | Integrin cell surface interactions, ECM proteoglycans, Lysosphingolipid and LPA receptors, Interleukin-4 and Interleukin-13 signaling |

**Supplementary Table 8. Stability selection to identify important Reactome pathways for drug response prediction in LEAP (PS-LASSO).** For each of the top 20 best predicted compounds in task 2, we report the prediction performance (Spearman's correlation) obtained with LEAP, the normalised stability score obtained with the tuned models, and the number and list of stably selected Reactome pathways (annotated with their identifiers or pathway names).

| Compound | Prediction performance with LEAP (PS-LASSO) | Stability score | Number of selected pathways | Selected pathway identifiers | Selected pathway names |
|---|---|---|---|---|---|
| necrosulfonamide | 0.717 | 0.86 | 16 | R-HSA-1251932, R-HSA-1300645, R-HSA-193807, R-HSA-2219530, R-HSA-3000157, R-HSA-3299685, R-HSA-5626978, R-HSA-6783783, R-HSA-6785807, R-HSA-8939246, R-HSA-8952158, R-HSA-9616222, R-HSA-975634, R-HSA-9818027, R-HSA-9820960, R-HSA-9827857 | PLCG1 events in ERBB2 signaling, Acrosome Reaction and Sperm:Oocyte Membrane Binding, Synthesis of bile acids and bile salts via 27-hydroxycholesterol, Constitutive Signaling by Aberrant PI3K in Cancer, Laminin interactions, Detoxification of Reactive Oxygen Species, TNFR1-mediated ceramide production, Interleukin-10 signaling, Interleukin-4 and Interleukin-13 signaling, RUNX1 regulates transcription of genes involved in differentiation of myeloid cells, RUNX3 regulates BCL2L11 (BIM) transcription, Transcriptional regulation of granulopoiesis, Retinoid metabolism and transport, NFE2L2 regulating anti-oxidant/detoxification enzymes, Respiratory syncytial virus (RSV) attachment and entry, Specification of primordial germ cells |
| refametinib | 0.714 | 0.86 | 35 | R-HSA-1296052, R-HSA-1592389, R-HSA-167826, R-HSA-167827, R-HSA-177135, R-HSA-190704, R-HSA-190827, R-HSA-192814, R-HSA-2160916, R-HSA-2485179, R-HSA-2892245, R-HSA-351200, R-HSA-3645790, R-HSA-3656248, R-HSA-418889, R-HSA-418890, R-HSA-434313, R-HSA-4419969, R-HSA-5579007, R-HSA-5601884, R-HSA-5619036, R-HSA-5662702, R-HSA-6804759, R-HSA-73728, R-HSA-75205, R-HSA-75876, R-HSA-8941413, R-HSA-8981607, R-HSA-9014843, R-HSA-9640463, R-HSA-9652817, R-HSA-9660826, R-HSA-9686347, R-HSA-9823730, R-HSA-9854909 | Ca2+ activated K+ channels, Activation of Matrix Metalloproteinases, The fatty acid cycling model, The proton buffering model, Conjugation of benzoate with glycine, Oligomerization of connexins into connexons, Transport of connexins along the secretory pathway, vRNA Synthesis, Hyaluronan uptake and degradation, Activation of the phototransduction cascade, POU5F1 (OCT4), SOX2, NANOG repress genes related to differentiation, Interconversion of polyamines, TGFBR2 Kinase Domain Mutants in Cancer, Defective HEXB causes GM2G2, Caspase activation via Dependence Receptors in the absence of ligand, Role of second messengers in netrin-1 signaling, Intracellular metabolism of fatty acids regulates insulin secretion, Depolymerization of the Nuclear Lamina, Defective ACY1 causes encephalopathy, PIWI-interacting RNA (piRNA) biogenesis, Defective SLC24A5 causes oculocutaneous albinism 6 (OCA6), Melanin biosynthesis, Regulation of TP53 Activity through Association with Co-factors, RNA Polymerase I Promoter Opening, Dissolution of Fibrin Clot, Synthesis of very long-chain fatty acyl-CoAs, Events associated with phagocytolytic activity of PMN cells, Intracellular oxygen transport, Interleukin-33 signaling, Wax biosynthesis, Signaling by MAPK mutants, Purinergic signaling in leishmaniasis infection, Microbial modulation of RIPK1-mediated regulated necrosis, Formation of definitive endoderm, Regulation of MITF-M dependent genes involved in invasion |
| triazolothiadiazine | 0.704 | 0.85 | 11 | R-HSA-2022377, R-HSA-209822, R-HSA-380994, R-HSA-416550, R-HSA-5666185, R-HSA-844455, R-HSA-8866910, R-HSA-8939246, R-HSA-9706369, R-HSA-9820960, R-HSA-9823730 | Metabolism of Angiotensinogen to Angiotensins, Glycoprotein hormones, ATF4 activates genes in response to endoplasmic reticulum stress, Sema4D mediated inhibition of cell attachment and migration, RHO GTPases Activate Rhotekin and Rhophilins, The NLRP1 inflammasome, TFAP2 (AP-2) family regulates transcription of growth factors and their receptors, RUNX1 regulates transcription of genes involved in differentiation of myeloid cells, Negative regulation of FLT3, Respiratory syncytial virus (RSV) attachment and entry, Formation of definitive endoderm |
| px-12 | 0.702 | 0.88 | 13 | R-HSA-210991, R-HSA-3000157, R-HSA-5609977, R-HSA-73614, R-HSA-73621, R-HSA-8866910, R-HSA-936837, R-HSA-9616222, R-HSA-9617828, R-HSA-9706369, R-HSA-975634, R-HSA-9818027, R-HSA-9827857 | Basigin interactions, Laminin interactions, Defective GALE causes EDG, Pyrimidine salvage, Pyrimidine catabolism, TFAP2 (AP-2) family regulates transcription of growth factors and their receptors, Ion transport by P-type ATPases, Transcriptional regulation of granulopoiesis, FOXO-mediated transcription of cell cycle genes, Negative regulation of FLT3, Retinoid metabolism and transport, NFE2L2 regulating anti-oxidant/detoxification enzymes, Specification of primordial germ cells |
| gsk-461364 | 0.673 | 0.91 | 19 | R-HSA-2022377, R-HSA-2024096, R-HSA-209822, R-HSA-352238, R-HSA-380994, R-HSA-416550, R-HSA-450520, R-HSA-5619068, R-HSA-5666185, R-HSA-73817, R-HSA-77387, R-HSA-844455, R-HSA-8862803, R-HSA-8866910, R-HSA-8939246, R-HSA-936837, R-HSA-9692913, R-HSA-9706369, R-HSA-9820960 | Metabolism of Angiotensinogen to Angiotensins, HS-GAG degradation, Glycoprotein hormones, Breakdown of the nuclear lamina, ATF4 activates genes in response to endoplasmic reticulum stress, Sema4D mediated inhibition of cell attachment and migration, HuR (ELAVL1) binds and stabilizes mRNA, Defective SLC2A10 causes arterial tortuosity syndrome (ATS), RHO GTPases Activate Rhotekin and Rhophilins, Purine ribonucleoside monophosphate biosynthesis, Insulin receptor recycling, The NLRP1 inflammasome, Deregulated CDK5 triggers multiple neurodegenerative pathways in Alzheimer's disease models, TFAP2 (AP-2) family regulates transcription of growth factors and their receptors, RUNX1 regulates transcription of genes involved in differentiation of myeloid cells, Ion transport by P-type ATPases, SARS-CoV-1-mediated effects on programmed cell death, Negative regulation of FLT3, Respiratory syncytial virus (RSV) attachment and entry |
| sb-225002 | 0.671 | 0.91 | 20 | R-HSA-1483166, R-HSA-1483191, R-HSA-189483, R-HSA-2022377, R-HSA-2408508, R-HSA-2892245, R-HSA-380994, R-HSA-416550, R-HSA-426048, R-HSA-450520, R-HSA-5221030, R-HSA-5666185, R-HSA-844455, R-HSA-8875513, R-HSA-8939246, R-HSA-936837, R-HSA-9692913, R-HSA-9706369, R-HSA-9758920, R-HSA-9823730 | Synthesis of PA, Synthesis of PC, Heme degradation, Metabolism of Angiotensinogen to Angiotensins, Metabolism of ingested SeMet, Sec, MeSec into H2Se, POU5F1 (OCT4), SOX2, NANOG repress genes related to differentiation, ATF4 activates genes in response to endoplasmic reticulum stress, Sema4D mediated inhibition of cell attachment and migration, Arachidonate production from DAG, HuR (ELAVL1) binds and stabilizes mRNA, TET1,2,3 and TDG demethylate DNA, RHO GTPases Activate Rhotekin and Rhophilins, The NLRP1 inflammasome, MET interacts with TNS proteins, RUNX1 regulates transcription of genes involved in differentiation of myeloid cells, Ion transport by P-type ATPases, SARS-CoV-1-mediated effects on programmed cell death, Negative regulation of FLT3, Formation of lateral plate mesoderm, Formation of definitive endoderm |
| sch772984 | 0.668 | 0.83 | 46 | R-HSA-1296052, R-HSA-1296067, R-HSA-1592389, R-HSA-167826, R-HSA-167827, R-HSA-177135, R-HSA-180689, R-HSA-190374, R-HSA-190704, R-HSA-190827, R-HSA-2046105, R-HSA-2142850, R-HSA-2160916, R-HSA-2485179, R-HSA-351200, R-HSA-352238, R-HSA-418889, R-HSA-418890, R-HSA-426117, R-HSA-434313, R-HSA-4419969, R-HSA-5339717, R-HSA-5576890, R-HSA-5579007, R-HSA-5619036, R-HSA-5619039, R-HSA-5619066, R-HSA-5632968, R-HSA-5652227, R-HSA-5662702, R-HSA-879518, R-HSA-8849473, R-HSA-8939247, R-HSA-8941413, R-HSA-8964046, R-HSA-8981607, R-HSA-9014843, R-HSA-9034013, R-HSA-9640463, R-HSA-9652817, R-HSA-9686347, R-HSA-9758881, R-HSA-9764302, R-HSA-9823730, R-HSA-9845620, R-HSA-9846298 | Ca2+ activated K+ channels, Potassium transport channels, Activation of Matrix Metalloproteinases, The fatty acid cycling model, The proton buffering model, Conjugation of benzoate with glycine, APOBEC3G mediated resistance to HIV-1 infection, FGFR1c and Klotho ligand binding and activation, Oligomerization of connexins into connexons, Transport of connexins along the secretory pathway, Linoleic acid (LA) metabolism, Hyaluronan biosynthesis and export, Hyaluronan uptake and degradation, Activation of the phototransduction cascade, Interconversion of polyamines, Breakdown of the nuclear lamina, Caspase activation via Dependence Receptors in the absence of ligand, Role of second messengers in netrin-1 signaling, Cation-coupled Chloride cotransporters, Intracellular metabolism of fatty acids regulates insulin secretion, Depolymerization of the Nuclear Lamina, Signaling by LRP5 mutants, Phase 3 - rapid repolarisation, Defective ACY1 causes encephalopathy, Defective SLC24A5 causes oculocutaneous albinism 6 (OCA6), Defective SLC12A6 causes agenesis of the corpus callosum, with peripheral neuropathy (ACCPN), Defective SLC22A18 causes lung cancer (LNCR) and embryonal rhabdomyosarcoma 1 (RMSE1), Defective Mismatch Repair Associated With MSH6, Fructose biosynthesis, Melanin biosynthesis, Transport of organic anions, PTK6 Expression, RUNX1 regulates transcription of genes involved in interleukin signaling, Events associated with phagocytolytic activity of PMN cells, VLDL clearance, Intracellular oxygen transport, Interleukin-33 signaling, NTF3 activates NTRK3 signaling, Wax biosynthesis, Signaling by MAPK mutants, Microbial modulation of RIPK1-mediated regulated necrosis, Uptake of dietary cobalamins into enterocytes, Regulation of CDH19 Expression and Function, Formation of definitive endoderm, Enhanced binding of GP1BA variant to VWF multimer:collagen, Defective binding of VWF variant to GPIb:IX:V |
| brd-k34222889 | 0.659 | 0.90 | 20 | R-HSA-111453, R-HSA-1266695, R-HSA-1433559, R-HSA-164944, R-HSA-2022377, R-HSA-210991, R-HSA-2219530, R-HSA-352238, R-HSA-416550, R-HSA-8866907, R-HSA-8939246, R-HSA-936837, R-HSA-9616222, R-HSA-9680350, R-HSA-9706369, R-HSA-975577, R-HSA-975634, R-HSA-9818027, R-HSA-9827857, R-HSA-9844594 | BH3-only proteins associate with and inactivate anti-apoptotic BCL-2 members, Interleukin-7 signaling, Regulation of KIT signaling, Nef and signal transduction, Metabolism of Angiotensinogen to Angiotensins, Basigin interactions, Constitutive Signaling by Aberrant PI3K in Cancer, Breakdown of the nuclear lamina, Sema4D mediated inhibition of cell attachment and migration, Activation of the TFAP2 (AP-2) family of transcription factors, RUNX1 regulates transcription of genes involved in differentiation of myeloid cells, Ion transport by P-type ATPases, Transcriptional regulation of granulopoiesis, Signaling by CSF1 (M-CSF) in myeloid cells, Negative regulation of FLT3, N-Glycan antennae elongation, Retinoid metabolism and transport, NFE2L2 regulating anti-oxidant/detoxification enzymes, Specification of primordial germ cells, Transcriptional regulation of brown and beige adipocyte differentiation by EBF2 |
| prima-1 | 0.659 | 0.88 | 19 | R-HSA-164944, R-HSA-189483, R-HSA-210991, R-HSA-2219530, R-HSA-351200, R-HSA-417973, R-HSA-5609977, R-HSA-73614, R-HSA-8851907, R-HSA-8939245, R-HSA-936837, R-HSA-9617828, R-HSA-9634638, R-HSA-9664565, R-HSA-975634, R-HSA-9818025, R-HSA-9818027, R-HSA-9823730, R-HSA-9827857 | Nef and signal transduction, Heme degradation, Basigin interactions, Constitutive Signaling by Aberrant PI3K in Cancer, Interconversion of polyamines, Adenosine P1 receptors, Defective GALE causes EDG, Pyrimidine salvage, MET activates PI3K/AKT signaling, RUNX1 regulates transcription of genes involved in BCR signaling, Ion transport by P-type ATPases, FOXO-mediated transcription of cell cycle genes, Estrogen-dependent nuclear events downstream of ESR-membrane signaling, Signaling by ERBB2 KD Mutants, Retinoid metabolism and transport, NFE2L2 regulating TCA cycle genes, NFE2L2 regulating anti-oxidant/detoxification enzymes, Formation of definitive endoderm, Specification of primordial germ cells |
| rigosertib | 0.657 | 0.82 | 44 | R-HSA-1306955, R-HSA-1483166, R-HSA-1483196, R-HSA-1614603, R-HSA-173107, R-HSA-176034, R-HSA-187024, R-HSA-187042, R-HSA-2022377, R-HSA-209822, R-HSA-2206302, R-HSA-2408508, R-HSA-2892245, R-HSA-3322077, R-HSA-3371598, R-HSA-390247, R-HSA-390696, R-HSA-416550, R-HSA-426048, R-HSA-5221030, R-HSA-5578996, R-HSA-5579014, R-HSA-5619066, R-HSA-5619068, R-HSA-5666185, R-HSA-5668599, R-HSA-6803544, R-HSA-73817, R-HSA-77387, R-HSA-844455, R-HSA-8866910, R-HSA-8939246, R-HSA-8949275, R-HSA-9027277, R-HSA-9615017, R-HSA-9673768, R-HSA-9692913, R-HSA-9706369, R-HSA-9733709, R-HSA-9734091, R-HSA-975577, R-HSA-9821993, R-HSA-9830364, R-HSA-9854909 | GRB7 events in ERBB2 signaling, Synthesis of PA, PI and PC transport between ER and Golgi membranes, Cysteine formation from homocysteine, Binding and entry of HIV virion, Interactions of Tat with host cellular proteins, NGF-independant TRKA activation, TRKA activation by NGF, Metabolism of Angiotensinogen to Angiotensins, Glycoprotein hormones, MPS I - Hurler syndrome, Metabolism of ingested SeMet, Sec, MeSec into H2Se, POU5F1 (OCT4), SOX2, NANOG repress genes related to differentiation, Glycogen synthesis, Defective BTD causes biotidinase deficiency, Beta-oxidation of very long chain fatty acids, Adrenoceptors, Sema4D mediated inhibition of cell attachment and migration, Arachidonate production from DAG, TET1,2,3 and TDG demethylate DNA, Defective CYP27A1 causes CTX, Defective CYP27B1 causes VDDR1A, Defective SLC22A18 causes lung cancer (LNCR) and embryonal rhabdomyosarcoma 1 (RMSE1), Defective SLC2A10 causes arterial tortuosity syndrome (ATS), RHO GTPases Activate Rhotekin and Rhophilins, RHO GTPases Activate NADPH Oxidases, Ion influx/efflux at host-pathogen interface, Purine ribonucleoside monophosphate biosynthesis, Insulin receptor recycling, The NLRP1 inflammasome, TFAP2 (AP-2) family regulates transcription of growth factors and their receptors, RUNX1 regulates transcription of genes involved in differentiation of myeloid cells, RUNX3 Regulates Immune Response and Cell Migration, Erythropoietin activates Phospholipase C gamma (PLCG), FOXO-mediated transcription of oxidative stress, metabolic and neuronal genes, Signaling by membrane-tethered fusions of PDGFRA or PDGFRB, SARS-CoV-1-mediated effects on programmed cell death, Negative regulation of FLT3, Cardiogenesis, Drug-mediated inhibition of MET activation, N-Glycan antennae elongation, Replacement of protamines by nucleosomes in the male pronucleus, Formation of the nephric duct, Regulation of MITF-M dependent genes involved in invasion |
| austocystin d | 0.656 | 0.81 | 29 | R-HSA-1306955, R-HSA-1433617, R-HSA-174403, R-HSA-190372, R-HSA-193634, R-HSA-196819, R-HSA-196836, R-HSA-210455, R-HSA-211935, R-HSA-3645790, R-HSA-390696, R-HSA-419812, R-HSA-420029, R-HSA-5340588, R-HSA-5576890, R-HSA-5579005, R-HSA-5579013, R-HSA-5602680, R-HSA-5652227, R-HSA-5656364, R-HSA-5657562, R-HSA-71032, R-HSA-71288, R-HSA-804914, R-HSA-8981373, R-HSA-8981607, R-HSA-9037629, R-HSA-9652817, R-HSA-9758919 | GRB7 events in ERBB2 signaling, Regulation of signaling by NODAL, Glutathione synthesis and recycling, FGFR3c ligand binding and activation, Axonal growth inhibition (RHOA activation), Vitamin B1 (thiamin) metabolism, Vitamin C (ascorbate) metabolism, Astrocytic Glutamate-Glutamine Uptake And Metabolism, Fatty acids, TGFBR2 Kinase Domain Mutants in Cancer, Adrenoceptors, Calcitonin-like ligand receptors, Tight junction interactions, Signaling by RNF43 mutants, Phase 3 - rapid repolarisation, Defective CYP4F22 causes ARCI5, Defective CYP7B1 causes SPG5A and CBAS3, MyD88 deficiency (TLR5), Fructose biosynthesis, Defective SLC5A1 causes congenital glucose/galactose malabsorption (GGM), Essential fructosuria, Propionyl-CoA catabolism, Creatine metabolism, Transport of fatty acids, Intestinal hexose absorption, Intracellular oxygen transport, Lewis blood group biosynthesis, Signaling by MAPK mutants, Epithelial-Mesenchymal Transition (EMT) during gastrulation |
| kw-2449 | 0.656 | 0.86 | 8 | R-HSA-2219530, R-HSA-449836, R-HSA-5666185, R-HSA-8866910, R-HSA-8939246, R-HSA-936837, R-HSA-9616222, R-HSA-9706369 | Constitutive Signaling by Aberrant PI3K in Cancer, Other interleukin signaling, RHO GTPases Activate Rhotekin and Rhophilins, TFAP2 (AP-2) family regulates transcription of growth factors and their receptors, RUNX1 regulates transcription of genes involved in differentiation of myeloid cells, Ion transport by P-type ATPases, Transcriptional regulation of granulopoiesis, Negative regulation of FLT3 |
| parbendazole | 0.653 | 0.89 | 7 | R-HSA-5666185, R-HSA-844455, R-HSA-8875513, R-HSA-8939246, R-HSA-936837, R-HSA-9706369, R-HSA-9820960 | RHO GTPases Activate Rhotekin and Rhophilins, The NLRP1 inflammasome, MET interacts with TNS proteins, RUNX1 regulates transcription of genes involved in differentiation of myeloid cells, Ion transport by P-type ATPases, Negative regulation of FLT3, Respiratory syncytial virus (RSV) attachment and entry |
| narciclasine | 0.647 | 0.87 | 16 | R-HSA-2022377, R-HSA-2024096, R-HSA-2219530, R-HSA-2408508, R-HSA-3299685, R-HSA-380994, R-HSA-416550, R-HSA-449836, R-HSA-5619068, R-HSA-844455, R-HSA-8939246, R-HSA-8957275, R-HSA-9616222, R-HSA-9706369, R-HSA-9761174, R-HSA-9820960 | Metabolism of Angiotensinogen to Angiotensins, HS-GAG degradation, Constitutive Signaling by Aberrant PI3K in Cancer, Metabolism of ingested SeMet, Sec, MeSec into H2Se, Detoxification of Reactive Oxygen Species, ATF4 activates genes in response to endoplasmic reticulum stress, Sema4D mediated inhibition of cell attachment and migration, Other interleukin signaling, Defective SLC2A10 causes arterial tortuosity syndrome (ATS), The NLRP1 inflammasome, RUNX1 regulates transcription of genes involved in differentiation of myeloid cells, Post-translational protein phosphorylation, Transcriptional regulation of granulopoiesis, Negative regulation of FLT3, Formation of intermediate mesoderm, Respiratory syncytial virus (RSV) attachment and entry |
| ly-2183240 | 0.646 | 0.88 | 8 | R-HSA-8866910, R-HSA-8875513, R-HSA-8939246, R-HSA-936837, R-HSA-9665686, R-HSA-9706369, R-HSA-9820960, R-HSA-9823730 | TFAP2 (AP-2) family regulates transcription of growth factors and their receptors, MET interacts with TNS proteins, RUNX1 regulates transcription of genes involved in differentiation of myeloid cells, Ion transport by P-type ATPases, Signaling by ERBB2 TMD/JMD mutants, Negative regulation of FLT3, Respiratory syncytial virus (RSV) attachment and entry, Formation of definitive endoderm |
| acy-1215 | 0.644 | 0.80 | 7 | R-HSA-111453, R-HSA-210991, R-HSA-416550, R-HSA-936837, R-HSA-9616222, R-HSA-9634638, R-HSA-9818027 | BH3-only proteins associate with and inactivate anti-apoptotic BCL-2 members, Basigin interactions, Sema4D mediated inhibition of cell attachment and migration, Ion transport by P-type ATPases, Transcriptional regulation of granulopoiesis, Estrogen-dependent nuclear events downstream of ESR-membrane signaling, NFE2L2 regulating anti-oxidant/detoxification enzymes |
| cr-1-31b | 0.640 | 0.84 | 63 | R-HSA-114516, R-HSA-114608, R-HSA-1236973, R-HSA-1251932, R-HSA-1300645, R-HSA-140834, R-HSA-1433559, R-HSA-1483115, R-HSA-1606341, R-HSA-174362, R-HSA-174430, R-HSA-1839120, R-HSA-187024, R-HSA-2022377, R-HSA-2024096, R-HSA-209822, R-HSA-2206302, R-HSA-2408508, R-HSA-3299685, R-HSA-3560783, R-HSA-3560796, R-HSA-380994, R-HSA-389887, R-HSA-416550, R-HSA-416572, R-HSA-425381, R-HSA-426048, R-HSA-5221030, R-HSA-5578996, R-HSA-5579014, R-HSA-5619039, R-HSA-5619049, R-HSA-5619054, R-HSA-5619060, R-HSA-5619068, R-HSA-5626978, R-HSA-5666185, R-HSA-5676594, R-HSA-5676934, R-HSA-5679001, R-HSA-629594, R-HSA-6787450, R-HSA-74217, R-HSA-74713, R-HSA-77108, R-HSA-844455, R-HSA-844623, R-HSA-8939246, R-HSA-8949275, R-HSA-8963684, R-HSA-9020265, R-HSA-9020933, R-HSA-9027277, R-HSA-9619665, R-HSA-9673768, R-HSA-9706369, R-HSA-9708296, R-HSA-9757110, R-HSA-9761174, R-HSA-9821993, R-HSA-9825895, R-HSA-9827857, R-HSA-9854909 | Disinhibition of SNARE formation, Platelet degranulation, Cross-presentation of particulate exogenous antigens (phagosomes), PLCG1 events in ERBB2 signaling, Acrosome Reaction and Sperm:Oocyte Membrane Binding, Extrinsic Pathway of Fibrin Clot Formation, Regulation of KIT signaling, Hydrolysis of LPC, IRF3 mediated activation of type 1 IFN, Transport and synthesis of PAPS, Telomere C-strand synthesis initiation, Signaling by FGFR1 amplification mutants, NGF-independant TRKA activation, Metabolism of Angiotensinogen to Angiotensins, HS-GAG degradation, Glycoprotein hormones, MPS I - Hurler syndrome, Metabolism of ingested SeMet, Sec, MeSec into H2Se, Detoxification of Reactive Oxygen Species, Defective B4GALT7 causes EDS, progeroid type, Defective PAPSS2 causes SEMD-PA, ATF4 activates genes in response to endoplasmic reticulum stress, Beta-oxidation of pristanoyl-CoA, Sema4D mediated inhibition of cell attachment and migration, Sema4D induced cell migration and growth-cone collapse, Bicarbonate transporters, Arachidonate production from DAG, TET1,2,3 and TDG demethylate DNA, Defective CYP27A1 causes CTX, Defective CYP27B1 causes VDDR1A, Defective SLC12A6 causes agenesis of the corpus callosum, with peripheral neuropathy (ACCPN), Defective SLC40A1 causes hemochromatosis 4 (HFE4) (macrophages), Defective SLC4A4 causes renal tubular acidosis, proximal, with ocular abnormalities and mental retardation (pRTA-OA), Defective CP causes aceruloplasminemia (ACERULOP), Defective SLC2A10 causes arterial tortuosity syndrome (ATS), TNFR1-mediated ceramide production, RHO GTPases Activate Rhotekin and Rhophilins, TNF receptor superfamily (TNFSF) members mediating non-canonical NF-kB pathway, Protein repair, Defective ABCC2 causes DJS, Highly calcium permeable postsynaptic nicotinic acetylcholine receptors, tRNA modification in the mitochondrion, Purine salvage, IRS activation, Utilization of Ketone Bodies, The NLRP1 inflammasome, The IPAF inflammasome, RUNX1 regulates transcription of genes involved in differentiation of myeloid cells, RUNX3 Regulates Immune Response and Cell Migration, Tyrosine catabolism, Biosynthesis of aspirin-triggered D-series resolvins, Interleukin-23 signaling, Erythropoietin activates Phospholipase C gamma (PLCG), EGR2 and SOX10-mediated initiation of Schwann cell myelination, Signaling by membrane-tethered fusions of PDGFRA or PDGFRB, Negative regulation of FLT3, tRNA-derived small RNA (tsRNA or tRNA-related fragment, tRF) biogenesis, Prednisone ADME, Formation of intermediate mesoderm, Replacement of protamines by nucleosomes in the male pronucleus, Regulation of MITF-M-dependent genes involved in DNA damage repair and senescence, Specification of primordial germ cells, Regulation of MITF-M dependent genes involved in invasion |
| (5z)-7-oxozeaenol | 0.639 | 0.89 | 11 | R-HSA-1296052, R-HSA-1592389, R-HSA-174403, R-HSA-193775, R-HSA-196836, R-HSA-442982, R-HSA-5578999, R-HSA-5619036, R-HSA-8981607, R-HSA-9652817, R-HSA-9764302 | Ca2+ activated K+ channels, Activation of Matrix Metalloproteinases, Glutathione synthesis and recycling, Synthesis of bile acids and bile salts via 24-hydroxycholesterol, Vitamin C (ascorbate) metabolism, Ras activation upon Ca2+ influx through NMDA receptor, Defective GCLC causes HAGGSD, Defective SLC24A5 causes oculocutaneous albinism 6 (OCA6), Intracellular oxygen transport, Signaling by MAPK mutants, Regulation of CDH19 Expression and Function |
| kx2-391 | 0.637 | 0.91 | 16 | R-HSA-2022377, R-HSA-2024096, R-HSA-209822, R-HSA-2408508, R-HSA-380994, R-HSA-416550, R-HSA-426048, R-HSA-446343, R-HSA-5619066, R-HSA-5666185, R-HSA-844455, R-HSA-8875513, R-HSA-8939246, R-HSA-9706369, R-HSA-9820960, R-HSA-9823730 | Metabolism of Angiotensinogen to Angiotensins, HS-GAG degradation, Glycoprotein hormones, Metabolism of ingested SeMet, Sec, MeSec into H2Se, ATF4 activates genes in response to endoplasmic reticulum stress, Sema4D mediated inhibition of cell attachment and migration, Arachidonate production from DAG, Localization of the PINCH-ILK-PARVIN complex to focal adhesions, Defective SLC22A18 causes lung cancer (LNCR) and embryonal rhabdomyosarcoma 1 (RMSE1), RHO GTPases Activate Rhotekin and Rhophilins, The NLRP1 inflammasome, MET interacts with TNS proteins, RUNX1 regulates transcription of genes involved in differentiation of myeloid cells, Negative regulation of FLT3, Respiratory syncytial virus (RSV) attachment and entry, Formation of definitive endoderm |
| sb-743921 | 0.630 | 0.84 | 9 | R-HSA-2022377, R-HSA-2024096, R-HSA-416550, R-HSA-446343, R-HSA-844455, R-HSA-8875513, R-HSA-8939246, R-HSA-9706369, R-HSA-9820960 | Metabolism of Angiotensinogen to Angiotensins, HS-GAG degradation, Sema4D mediated inhibition of cell attachment and migration, Localization of the PINCH-ILK-PARVIN complex to focal adhesions, The NLRP1 inflammasome, MET interacts with TNS proteins, RUNX1 regulates transcription of genes involved in differentiation of myeloid cells, Negative regulation of FLT3, Respiratory syncytial virus (RSV) attachment and entry |
| clofarabine | 0.630 | 0.88 | 14 | R-HSA-114604, R-HSA-1855204, R-HSA-198933, R-HSA-2024096, R-HSA-2219530, R-HSA-6811558, R-HSA-8939246, R-HSA-8949275, R-HSA-9013149, R-HSA-9616222, R-HSA-9705677, R-HSA-9706369, R-HSA-975634, R-HSA-9820960 | GPVI-mediated activation cascade, Synthesis of IP3 and IP4 in the cytosol, Immunoregulatory interactions between a Lymphoid and a non-Lymphoid cell, HS-GAG degradation, Constitutive Signaling by Aberrant PI3K in Cancer, PI5P, PP2A and IER3 Regulate PI3K/AKT Signaling, RUNX1 regulates transcription of genes involved in differentiation of myeloid cells, RUNX3 Regulates Immune Response and Cell Migration, RAC1 GTPase cycle, Transcriptional regulation of granulopoiesis, SARS-CoV-2 targets PDZ proteins in cell-cell junction, Negative regulation of FLT3, Retinoid metabolism and transport, Respiratory syncytial virus (RSV) attachment and entry |

**Supplementary Table 9: Human assessment of biological plausibility of important Reactome pathways for top predicted drugs.** For each drug, we report the compound name, known mechanism of action, prediction performance with LEAP (PS-LASSO), list of selected pathways and highlighted pathways with plausible biological implications.

| Compound | Mechanism of action | Prediction performance with LEAP (PS-LASSO) | Selected pathway names | Human review of plausibility |
|---|---|---|---|---|
| necrosulfonamide | Necrosulfonamide is an inhibitor of necroptosis, a form of programmed cell death. It specifically targets the mixed lineage kinase domain-like protein (MLKL), preventing it from interacting with its downstream effectors and halting the necroptotic process. | 0.717 | **PLCG1 events in ERBB2 signaling,** Acrosome Reaction and Sperm:Oocyte Membrane Binding, Synthesis of bile acids and bile salts via 27-hydroxycholesterol, Constitutive Signaling by Aberrant PI3K in Cancer, Laminin interactions, **Detoxification of Reactive Oxygen Species**, **TNFR1-mediated ceramide production**, Interleukin-10 signaling, Interleukin-4 and Interleukin-13 signaling, RUNX1 regulates transcription of genes involved in differentiation of myeloid cells, RUNX3 regulates BCL2L11 (BIM) transcription, Transcriptional regulation of granulopoiesis, Retinoid metabolism and transport, **NFE2L2 regulating anti-oxidant/detoxification enzymes**, Respiratory syncytial virus (RSV) attachment and entry, Specification of primordial germ cells | **TNFR1-mediated ceramide production**: The production of ceramide is a downstream event of Tumor necrosis factor receptor 1 (TNFR1) activation and is involved in cell death signaling. TNFR1 signaling can trigger necroptosis. By inhibiting MLKL, necrosulfonamide blocks a key step in this pathway.<br>**Detoxification of Reactive Oxygen Species**, **NFE2L2 regulating anti-oxidant/detoxification enzymes**: necroptosis is often associated with the generation of ROS. By inhibiting necroptosis, necrosulfonamide could indirectly influence pathways related to ROS detoxification.<br>**PLCG1 events in ERBB2 signaling**: Phospholipase C gamma 1 is activated by extracellular stimuli and plays a crucial role in regulating various cellular functions necessary for tumorigenesis. Inhibition of PLCγ1 phosphorylation was associated to increased apoptosis and necroptosis. |
| refametinib | Refametinib is a selective inhibitor of MEK1 and MEK2, which are key components of the RAS/RAF/MEK/ERK signaling pathway that regulates cell growth, often constitutively activated in many cancers. | 0.714 | Ca2+ activated K+ channels, **Activation of Matrix Metalloproteinases**, The fatty acid cycling model, The proton buffering model, Conjugation of benzoate with glycine, Oligomerization of connexins into connexons, Transport of connexins along the secretory pathway, vRNA Synthesis, Hyaluronan uptake and degradation, Activation of the phototransduction cascade, **POU5F1 (OCT4), SOX2, NANOG repress genes related to differentiation**, Interconversion of polyamines, **TGFBR2 Kinase Domain Mutants in Cancer**, Defective HEXB causes GM2G2, Caspase activation via Dependence Receptors in the absence of ligand, Role of second messengers in netrin-1 signaling, Intracellular metabolism of fatty acids regulates insulin secretion, Depolymerization of the Nuclear Lamina, Defective ACY1 causes encephalopathy, PIWI-interacting RNA (piRNA) biogenesis, Defective SLC24A5 causes oculocutaneous albinism 6 (OCA6), **Melanin biosynthesis**, Regulation of TP53 Activity through Association with Co-factors, RNA Polymerase I Promoter Opening, Dissolution of Fibrin Clot, Synthesis of very long-chain fatty acyl-CoAs, Events associated with phagocytolytic activity of PMN cells, Intracellular oxygen transport, Interleukin-33 signaling, Wax biosynthesis, **Signaling by MAPK mutants**, Purinergic signaling in leishmaniasis infection, Microbial modulation of RIPK1-mediated regulated necrosis, **Formation of definitive endoderm**, **Regulation of MITF-M dependent genes involved in invasion** | **Signaling by MAPK mutants**: most direct association, as by inhibiting MEK1/2, refametinib directly targets a central component of the MAPK signaling pathway.<br>**Activation of Matrix Metalloproteinases**: The expression and activity of MMPs are often regulated by MAPK signaling.<br>**POU5F1 (OCT4), SOX2, NANOG repress genes related to differentiation**: These are key transcription factors in embryonic stem cells, and their expression can be regulated by signaling pathways like MAPK.<br>**TGFBR2 Kinase Domain Mutants in Cancer**: The TGF-beta signaling pathway can interact with the MAPK pathway.<br>**Regulation of MITF-M dependent genes involved in invasion**, **Melanin biosynthesis**: MITF is a key regulator of melanocyte development and melanoma, and its activity is modulated by the MAPK pathway.<br>**Formation of definitive endoderm**: MAPK pathways are able to regulate both the early embryonic development and the ES cell commitment from early steps of the process to mature differentiated cells.<br>**Pathways related to cell growth, differentiation, and metabolism:** The MAPK pathway plays a critical role in a vast array of cellular processes. Therefore, its inhibition by refametinib can plausibly affect. |
| triazolothiadiazine | Triazolothiadiazine derivatives are known to have a wide range of biological activities, including anticancer, antiviral, and anti-inflammatory properties. Some have been shown to induce apoptosis and cell cycle arrest in cancer cells. | 0.704 | Metabolism of Angiotensinogen to Angiotensins, Glycoprotein hormones, **ATF4 activates genes in response to endoplasmic reticulum stress**, Sema4D mediated inhibition of cell attachment and migration, RHO GTPases Activate Rhotekin and Rhophilins, **The NLRP1 inflammasome**, TFAP2 (AP-2) family regulates transcription of growth factors and their receptors, **RUNX1 regulates transcription of genes involved in differentiation of myeloid cells**, **Negative regulation of FLT3**, **Respiratory syncytial virus (RSV) attachment and entry**, **Formation of definitive endoderm** | The selected pathways are overall quite broad, potentially reflecting the broad spectrum of mechanisms for triazolothiadiazine and derivatives, including: 1) inhibition of various enzyme activities e.g., protein kinases, proteases, or hydrolases, leading to anti-inflammatory, anticancer, or antimicrobial effects. 2) Receptor modulations (eg. GABA, Adrenergic) 3) inhibition of inflammatory mediators (e.g., COX enzymes) or modulation of cytokine production 4) scavenging free radicals and reducing oxidative stress.<br>**ATF4 activates genes in response to endoplasmic reticulum stress**: ER stress can trigger apoptosis, a process that can be induced by some triazolothiadiazine derivatives.<br>**The NLRP1 inflammasome**: The NLRP1 inflammasome is involved in inflammation and pyroptosis, a form of programmed cell death. The anti-inflammatory properties of triazolothiadiazines suggest a potential interaction with this pathway.<br>**RUNX1 regulates transcription of genes involved in differentiation of myeloid cells**: Given that some triazolothiadiazines have anticancer effects, they may influence the differentiation of hematopoietic cells.<br>**Negative regulation of FLT3**: FLT3 is a receptor tyrosine kinase often mutated in acute myeloid leukemia. The anticancer properties of triazolothiadiazines could involve the modulation of such signaling pathways.<br>**Respiratory syncytial virus (RSV) attachment and entry**: The antiviral activity of some triazolothiadiazine derivatives could be due to interference with viral entry or replication processes.<br>**Formation of definitive endoderm**: As with other compounds affecting fundamental cellular processes, an influence on embryonic development is plausible though indirect. |
| px-12 | PX-12 is an inhibitor of thioredoxin-1 (Trx-1), a protein involved in redox regulation and cell growth. By inhibiting Trx-1, PX-12 can induce apoptosis and has shown anti-tumor activity. | 0.702 | Basigin interactions, **Laminin interactions**, Defective GALE causes EDG, Pyrimidine salvage, Pyrimidine catabolism, TFAP2 (AP-2) family regulates transcription of growth factors and their receptors, Ion transport by P-type ATPases, **Transcriptional regulation of granulopoiesis**, **FOXO-mediated transcription of cell cycle genes**, **Negative regulation of FLT3**, **Retinoid metabolism and transport**, **NFE2L2 regulating anti-oxidant/detoxification enzymes**, Specification of primordial germ cells | **FOXO-mediated transcription of cell cycle genes**: Thioredoxin-1 can regulate the activity of transcription factors. FOXO transcription factors are involved in cell cycle arrest and apoptosis, and their activity can be modulated by the cellular redox state, which is altered by PX-12.<br>**Negative regulation of FLT3**: FLT3 signaling could also be influenced by the redox environment.<br>**NFE2L2 regulating anti-oxidant/detoxification enzymes**: Trx-1 is a key component of the antioxidant system. Inhibition of Trx-1 by PX-12 would directly impact cellular redox balance and could, in turn, affect the NFE2L2-mediated antioxidant response.<br>**Transcriptional regulation of granulopoiesis**, **Specification of primordial germ cells**: The development and differentiation of cells are sensitive to the cellular redox state. By modulating Trx-1 activity, PX-12 could influence these developmental processes.<br>**Laminin interactions**, **Basigin interactions**: Cell adhesion and interactions with the extracellular matrix can be influenced by the redox state of cell surface proteins.<br>**Retinoid metabolism and transport**: Retinoid metabolism is a redox-dependent process, suggesting a potential link to the mechanism of action of PX-12. |
| gsk-461364 | GSK-461364 is a selective inhibitor of Polo-like kinase 1 (Plk1), a key regulator of the cell cycle, particularly during mitosis. Inhibition of Plk1 leads to mitotic arrest and cell death in cancer cells. | 0.673 | **Metabolism of Angiotensinogen to Angiotensins**, HS-GAG degradation, **Glycoprotein hormones**, **Breakdown of the nuclear lamina**, ATF4 activates genes in response to endoplasmic reticulum stress, Sema4D mediated inhibition of cell attachment and migration, HuR (ELAVL1) binds and stabilizes mRNA, Defective SLC2A10 causes arterial tortuosity syndrome (ATS), RHO GTPases Activate Rhotekin and Rhophilins, Purine ribonucleoside monophosphate biosynthesis, Insulin receptor recycling, The NLRP1 inflammasome, **Deregulated CDK5 triggers multiple neurodegenerative pathways in Alzheimer's disease models**, TFAP2 (AP-2) family regulates transcription of growth factors and their receptors, **RUNX1 regulates transcription of genes involved in differentiation of myeloid cells**, Ion transport by P-type ATPases, SARS-CoV-1-mediated effects on programmed cell death, Negative regulation of FLT3, **Respiratory syncytial virus (RSV) attachment and entry** | **Breakdown of the nuclear lamina**: Plk1 plays a crucial role in the disassembly of the nuclear lamina during mitosis.<br>**Deregulated CDK5 triggers multiple neurodegenerative pathways in Alzheimer's disease models**: CDK5 is another kinase involved in cell cycle regulation, and its deregulation is implicated in neurodegenerative diseases. There may be crosstalk between these kinase signaling pathways.<br>**RUNX1 regulates transcription of genes involved in differentiation of myeloid cells**: As Plk1 is essential for cell division, its inhibition would impact the proliferation and differentiation of hematopoietic cells.<br>**Respiratory syncytial virus (RSV) attachment and entry**: Some viruses manipulate the host cell cycle to facilitate their replication. By arresting the cell cycle, GSK-461364 could interfere with the RSV life cycle.<br>**Metabolism of Angiotensinogen to Angiotensins**, **Glycoprotein hormones**: While less direct, the production and signaling of hormones can be linked to the cell cycle and proliferation of endocrine cells. |
| sb-225002 | SB-225002 is a potent and selective antagonist of the CXCR2 chemokine receptor. CXCR2 is involved in the recruitment of neutrophils and other immune cells to sites of inflammation. | 0.671 | Synthesis of PA, Synthesis of PC, Heme degradation, Metabolism of Angiotensinogen to Angiotensins, Metabolism of ingested SeMet, Sec, MeSec into H2Se, POU5F1 (OCT4), SOX2, NANOG repress genes related to differentiation, ATF4 activates genes in response to endoplasmic reticulum stress, Sema4D mediated inhibition of cell attachment and migration, Arachidonate production from DAG, HuR (ELAVL1) binds and stabilizes mRNA, TET1,2,3 and TDG demethylate DNA, RHO GTPases Activate Rhotekin and Rhophilins, **The NLRP1 inflammasome**, MET interacts with TNS proteins, **RUNX1 regulates transcription of genes involved in differentiation of myeloid cells**, Ion transport by P-type ATPases, **SARS-CoV-1-mediated effects on programmed cell death**, **Negative regulation of FLT3**, Formation of lateral plate mesoderm, Formation of definitive endoderm | **The NLRP1 inflammasome**: The activation of the NLRP1 inflammasome leads to the production of inflammatory cytokines that can drive the recruitment of neutrophils via CXCR2.<br>**SARS-CoV-1-mediated effects on programmed cell death**: Viral infections, including those by coronaviruses, can trigger a strong inflammatory response characterized by the influx of neutrophils. Recent evidence supports the role of neutrophils in the pathogenesis of COVID-19 and proposed CXCR2 inhibition as a promising treatment option to block neutrophil recruitment and activation.<br>**Negative regulation of FLT3**: FLT3 is a cytokine receptor, there could be an indirect link.<br>**RUNX1 regulates transcription of genes involved in differentiation of myeloid cells**: Chemokine signaling is crucial for the proper development and trafficking of immune cells, including those regulated by the transcription factor RUNX1. |
| sch772984 | Sch772984 is a potent and specific inhibitor of ERK1 and ERK2, which are key kinases in the MAPK signaling pathway | 0.668 | Ca2+ activated K+ channels, Potassium transport channels, **Activation of Matrix Metalloproteinases**, The fatty acid cycling model, The proton buffering model, Conjugation of benzoate with glycine, APOBEC3G mediated resistance to HIV-1 infection, FGFR1c and Klotho ligand binding and activation, Oligomerization of connexins into connexons, Transport of connexins along the secretory pathway, Linoleic acid (LA) metabolism, Hyaluronan biosynthesis and export, Hyaluronan uptake and degradation, Activation of the phototransduction cascade, Interconversion of polyamines, **Breakdown of the nuclear lamina**, Caspase activation via Dependence Receptors in the absence of ligand, Role of second messengers in netrin-1 signaling, Cation-coupled Chloride cotransporters, Intracellular metabolism of fatty acids regulates insulin secretion, Depolymerization of the Nuclear Lamina, Signaling by LRP5 mutants, Phase 3 - rapid repolarisation, Defective ACY1 causes encephalopathy, Defective SLC24A5 causes oculocutaneous albinism 6 (OCA6), Defective SLC12A6 causes agenesis of the corpus callosum, with peripheral neuropathy (ACCPN), Defective SLC22A18 causes lung cancer (LNCR) and embryonal rhabdomyosarcoma 1 (RMSE1), Defective Mismatch Repair Associated With MSH6, Fructose biosynthesis, **Melanin biosynthesis**, Transport of organic anions, PTK6 Expression, RUNX1 regulates transcription of genes involved in interleukin | **Signaling by MAPK mutants**: most direct association. Sch772984 directly inhibits the core components of the MAPK pathway.<br>**Activation of Matrix Metalloproteinases**: The expression and activity of MMPs are often regulated by MAPK signaling.<br>**Breakdown of the nuclear lamina**: ERK signaling can influence the phosphorylation of lamins, which is necessary for nuclear envelope breakdown during mitosis.<br>**Melanin biosynthesis**: MITF activity is regulated by a direct interaction with RAF proteins in melanoma cells. As with refametinib, the inhibition of ERK could directly impact MITF. MITF regulates biogenesis of organelles such as the lysosome and endosomes through its regulation of components of the V-ATPase that is required for organelle acidification and function. MITF-dependent targets also contribute to autophagy when cells are stressed for nutrients. MITF additionally controls expression of the lysosome-resident acid ceramidase ASAH1 that has roles in sphingolipid metabolism and cellular proliferation in melanoma.<br>**Formation of definitive endoderm**: MAPK pathways are able to regulate both the early embryonic development and the ES cell commitment from early steps of the process to mature differentiated cells. |

| | | |
|---|---|---|
| | Expression, RUNX1 regulates transcription of genes involved in interleukin signaling, Events associated with phagocytolytic activity of PMN cells, VLDL clearance, Intracellular oxygen transport, **Interleukin-33 signaling**, NTF3 activates NTRK3 signaling, Wax biosynthesis, **Signaling by MAPK mutants**, Microbial modulation of RIPK1-mediated regulated necrosis, Uptake of dietary cobalamins into enterocytes, Regulation of CDH19 Expression and Function, | **Interleukin-33 signaling**: IL-33 can activate the MAPK pathway, so inhibiting ERK would block downstream signaling from this cytokine.<br><br>Pathways related to related to ion transport, metabolism, and cell adhesion: The extensive involvement of the MAPK pathway in cellular signaling provides a plausible, though often indirect, link to the other listed pathways. |